\documentclass{article}

\usepackage{microtype}
\usepackage{graphicx}
\usepackage{subfigure}
\usepackage{booktabs} %
\usepackage{caption}
\usepackage{subcaption}
\usepackage{hyperref}

\usepackage[accepted]{icml2025}

\usepackage{amsmath}
\usepackage{amssymb}
\usepackage{mathtools}
\usepackage{amsthm}

\usepackage[capitalize,noabbrev]{cleveref}

\usepackage{multicol}
\usepackage{multirow}

\theoremstyle{plain}

\theoremstyle{definition}

\theoremstyle{remark}

\usepackage[textsize=tiny]{todonotes}

\usepackage{graphicx}
\usepackage{wrapfig}
\usepackage[skins,breakable]{tcolorbox}
\usepackage{mdframed}%
\usepackage{xcolor}%
\usepackage{soul}
\usepackage{bbm}

\definecolor{light-purple}{RGB}{151,156,171}
\definecolor{blue-color}{RGB}{40,166,189}
\definecolor{pink-color}{RGB}{237,46,104} 
\definecolor{dark-grey-color}{RGB}{79,91,102}

\newcommand{\promptsubsection}[1]{
\setlength{\parskip}{6pt} \noindent\textbf{{#1}:}
}
\newcommand{\param}[1]{
\textcolor{pink-color}{\scriptsize{\texttt{\detokenize{#1}}}}
}

\newcommand{\prompttext}[1]{\textcolor{dark-grey-color}{#1}}

\newtcolorbox[list inside=prompt,auto counter,number within=section]{prompt}[1][]{
    colbacktitle=black!80,
    colframe=black!80,
    coltitle=white,
    fontupper=\footnotesize,
    boxsep=5pt,
    left=0pt,
    right=0pt,
    top=0pt,
    bottom=0pt,
    boxrule=1pt,
    enhanced, 
    breakable,
    skin first=enhanced,
    skin middle=enhanced,
    skin last=enhanced,
    #1,
}

\icmltitlerunning{Understanding Synthetic Context Extension via Retrieval Heads}

\begin{document}

\twocolumn[
\icmltitle{Understanding Synthetic Context Extension via Retrieval Heads}

\icmlsetsymbol{equal}{*}

\begin{icmlauthorlist}
\icmlauthor{Xinyu Zhao}{utaustin}
\icmlauthor{Fangcong Yin}{utaustin}
\icmlauthor{Greg Durrett}{utaustin}
\end{icmlauthorlist}

\icmlaffiliation{utaustin}{Department of Computer Science, The University of Texas at Austin, Texas, USA}

\icmlcorrespondingauthor{Xinyu Zhao}{xinyuzhao@utexas.edu}

\icmlkeywords{Machine Learning, ICML, Large Language Models, Synthetic Data, Retrieval Heads}

\vskip 0.3in
]

\printAffiliationsAndNotice{}  %

\begin{abstract}
Long-context LLMs are increasingly in demand for applications such as retrieval-augmented generation. To defray the cost of pretraining LLMs over long contexts, recent work takes an approach of synthetic context extension: fine-tuning LLMs with synthetically-generated long-context data. However, it remains unclear how and why this synthetic context extension imparts abilities for downstream long-context tasks. In this paper, we investigate fine-tuning on synthetic data for three long-context tasks that require retrieval and reasoning. We vary the realism of ``needle'' concepts to be retrieved and diversity of the surrounding ``haystack'' context, from using LLMs to construct synthetic documents to using templated relations and creating symbolic datasets. Although models trained on synthetic data underperform models trained on the real data, the impacts of both training settings can be understood via a shared feature of the attention computation, \textit{retrieval heads} \cite{wu2025retrieval}. The retrieval heads learned from synthetic data have high overlap with retrieval heads learned on real data. Furthermore, there is a strong correlation between the recall of heads learned and the downstream performance of a model, allowing us to interpret and predict the performance of models trained in different settings. Our results shed light on how to interpret synthetic data fine-tuning performance and how to approach creating better data for learning real-world LLM capabilities over long contexts.
\end{abstract}

\section{Introduction}

The quadratic memory scaling of Transformer attention imposes a strong computational constraint on our ability to train and do inference on long-context models. This disrupts the typical pre-training pipeline: pre-training must be done on as much data as possible, but pre-training a long context model would necessarily reduce the number of observed tokens able to fit on the GPU. One solution for this is to rely on synthetic data, now common in post-training settings such as SFT \citep{Xu2023WizardLMEL,yue2024mammoth,Xu2024MagpieAD,Chen2024GenQAGM} and RLHF/DPO \citep{Yang2023RLCDRL}. Recent prior work has proposed using synthetic data to extend the long-context abilities of LLMs after pre-training \citep{xiong2024artificial,Zhao2024LongSkyworkAT, llama3_1}.

This use of synthetic data is particularly necessary for long context tasks since they are so laborious for humans to manually label. Synthetic data is also configurable: it can exhibit different reasoning skills and ``teach'' models have to make certain types of inferences \citep{du-etal-2017-learning,wei2018fast,agarwal-etal-2021-knowledge,Tang-Et-Al:2024:MiniCheck,Divekar2024SynthesizRRGD}.
One way to do this is using templates to express pieces of information that must be reasoned over and to create symbolic tasks that are thought to mirror the reasoning required in the real task \citep{hsieh2024ruler,prakash2024fine,PrOntoQA,li2024needlebenchllmsretrievalreasoning}. However, past work has shown varying results from training on data for this kind of context extension \citep{fu2024data}; we lack general understanding of what properties are needed for good synthetic data.

In this paper, we explore several methods of creating synthetic long context data across three tasks. Our goal is to examine what makes synthetic data effective for this kind of context extension. While more realistic data is often better, it is unreliable: certain types of more synthetic data can exhibit desired long-context patterns even more effectively and with fewer shortcuts than realistic data. However, other types of synthetic data severely underperform on these tasks.

\begin{figure*}[t!]
\centering
\includegraphics[trim={0cm 7.5cm 4cm 3cm},clip,width=0.95\textwidth]{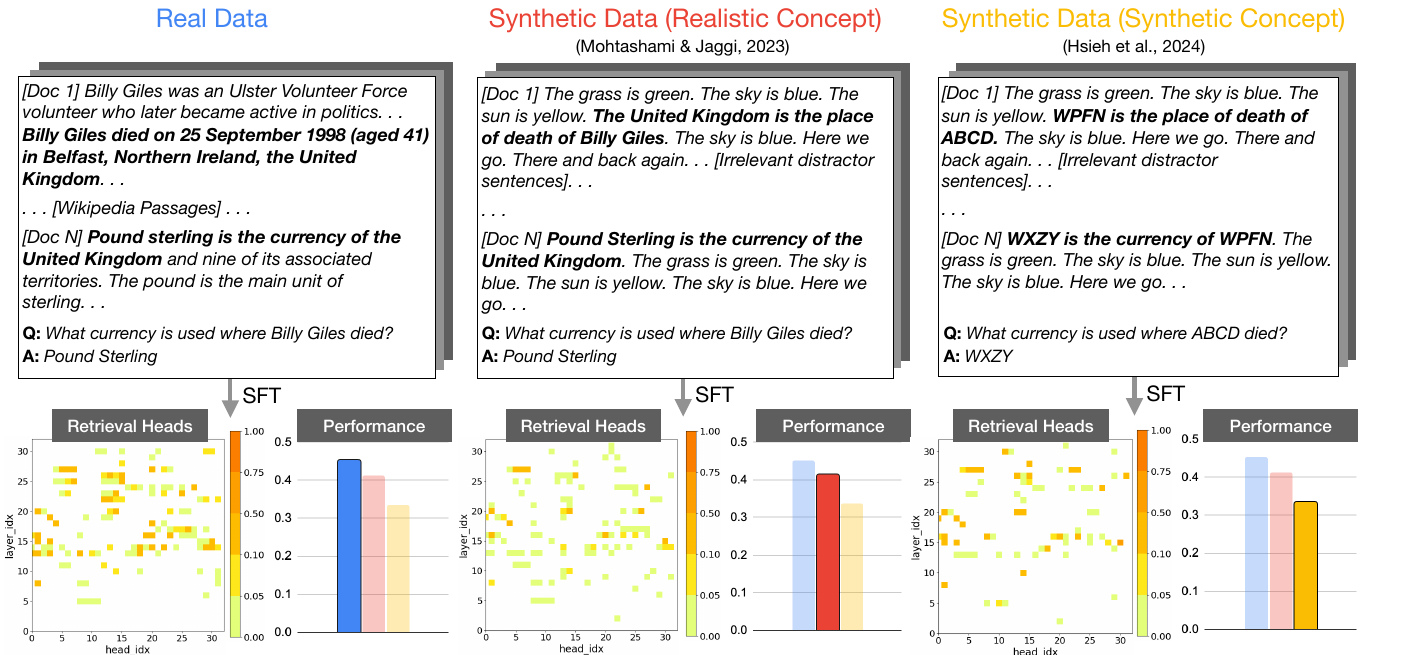}
\caption{We explore synthetic context extension with different forms of synthetic data across multiple tasks. Examples for a two-hop question from MuSiQue \citep{trivedi2021musique} are shown here. A special set of attention heads, \textit{retrieval heads} \citep{wu2025retrieval}, help explain the performance gap between fine-tuning on real data and synthetic data.}
\label{figure:title_figure}
\end{figure*}

To understand this divergence, we analyze models finetuned on different synthetic long-context datasets for the presence of a special set of attention heads called retrieval heads \citep{wu2025retrieval}.  %
Figure~\ref{figure:title_figure} shows two of our key results. First, the retrieval heads learned on poor-performing synthetic data tend to be fewer than those learned on realistic or high-quality synthetic data. Second, we find that the cosine similarity of retrieval scores of individual heads learned on synthetic data and realistic data correlates strongly with the downstream performance. Learning the ``right" set of retrieval heads seems to be a necessary condition for high performance. However, it is not sufficient: even when our synthetic data largely trains the \textit{same} retrieval heads required for the real task, it does so less effectively. We show that patching heads at the intersection of a poor-performing model and a high-performing model can improve performance of the former: these heads are where important operations are happening, but realistic data teaches them more strongly.
While these heads do not tell an entire story of the long-context understanding mechanisms, they serve as an effective indicator of the subnetworks affected during fine-tuning.

Our contributions are: (1) analysis of synthetic data across three synthetic tasks for long-context LLM training to determine the effect of varying data realism, including using symbolic data; (2) experimental results showing that learned retrieval head scores correlates with effectiveness of the training data for this setting. Taken together, we believe this work indicates a path forward for how to engineer better synthetic data and how to connect the construction process of synthetic data to (a) what it teaches Transformers and (b) how those models perform on downstream tasks.

\section{Background and Setup} 
\subsection{Background: Synthetic Data for Training LMs} 

Formally, consider a supervised learning setting for a pre-trained transformer language model $\mathcal{M}$. Given a task $\mathcal{T}$, we assume a distribution $p_{\mathcal{T}}$ of real-world task instances. We assume that a small, limited set of input-label pairs $\mathcal{D_{T}} = (x_{\mathcal T},y_{\mathcal T})$ drawn from the distribution $p_{\mathcal{T}}$ is available as seed data. A synthetic dataset $\tilde{\mathcal{D}}_\mathcal{T}$ is a set of input-label pairs sampled from the outputs of a data generator $\mathcal{G}$ given the seed data or the known properties: $\tilde{\mathcal{D}}_\mathcal{T} \sim p((\tilde{x},\tilde{y}) \mid \mathcal{D_{T}})$. Benchmarking or training $\mathcal{M}$ on such a synthetic $\tilde{\mathcal{D}}_\mathcal{T}$ is expected to evaluate or teach $\mathcal{M}$ the capabilities that can be \textit{transferred} to the real-world distribution $p_{\mathcal{T}}$.

A recent line of work has shown that simple heuristic-based synthetic datasets can be surprisingly effective for \textit{context extension}, a post-training scenario where LLMs that have been pre-trained on short-context corpora are further trained on long-context tasks to extend the effective context window \citep{fu2024data,Zhao2024LongSkyworkAT,xiong2024artificial}. For example, \citet{xiong2024artificial} finds that fine-tuning on a synthetic simple dictionary key-value retrieval task can even outperform models fine-tuned on realistic in-domain data. Other types of simplified and symbolic data are used in long-context benchmarks \cite{hsieh2024ruler, li2024needlebenchllmsretrievalreasoning}, despite the fact that the overlap between these and realistic data capabilities has not been thoroughly studied.

We call these approaches \textbf{synthetic context extension}: using synthetic data to extend the context window of LLMs. It remains unclear how and why synthetic data, especially when drawn from a very different distribution from the real data, can be effective despite results that support the contrary \citep{Chen2024UnveilingTF,Liu2024BestPA}. There is also a lack of general principles for creating synthetic training data beyond dataset-specific constructions in the literature. We start by constructing synthetic datasets varying in systematic ways to unify these variants from the literature.

\subsection{Experimental Setup}

Following \citet{xiong2024artificial}, we focus on fine-tuning LLMs for long-context retrieval and reasoning tasks where training on high-quality synthetic data has been shown to outperform real data. We also extend to multi-hop settings. We experiment on tasks where, given a long context $\mathcal{C}$ and a context-based query $q$, a language model $\mathcal{M}$ needs to retrieve one or more ``needle concepts'' $f_1,\ldots,f_m$ from $\mathcal{C}$ (pieces of relevant information), reason over that information, and then generate a response $\mathbf{\tilde{y}} \sim p( y \mid \mathcal{C},q)$ where $p(y \mid \mathcal{C},q)$ is the conditional distribution that $M$ places over the vocabulary $\Sigma^*$ given the context and the query. We consider extending the context window from 8K to 32K tokens to be representative of synthetic context extension following \citet{Chen2023ExtendingCW}. We use the following datasets.

\textbf{MDQA \citep{liu-etal-2024-lost}:} MDQA is a multi-document question answering (QA) dataset where only one paragraph in $C$ contains the gold answer to a single-hop query; that is, there is a single $f$ which directly addresses $q$. We extend the original MDQA dataset in 4K context to 32K context by retrieving additional distractor paragraphs from Natural Questions-Open \citep{kwiatkowski-etal-2019-natural, lee-etal-2019-latent} with Contriever \citep{izacard2021contriever}.

\textbf{MuSiQue \citep{trivedi2021musique}:} MuSiQue is a multi-hop QA dataset where the model must identify a piece of relevant information from a different document for each hop of the question in order to retrieve the final correct answer from the context. We use the linear three-hop subset of MuSiQue and extend the dataset to 32K by adding padding paragraphs to the original context.\footnote{Following \cite{mohtashami2023randomaccess}, we pad with irrelevant repeated text ``The grass is green. The sky is blue..." to ensure that the added paragraphs do not interfere with the answer to the original question.} In this setting, the facts $f_1,f_2,f_3$ are natural language sentences expressing knowledge triples.\footnote{Note that this is different from the two-hop examples in \mbox{Figure~\ref{figure:title_figure}} used for demonstrative purposes.}

\textbf{SummHay Citation \citep{laban2024summary}:} Summary of a Haystack (SummHay) is a long-context retrieval dataset where the model is given a set of documents with controlled ``insights,'' and asked to produce a list of key points. Additionally, the model must cite the correct documents in support of each key point. We isolate the citation component and construct a task where, given a haystack of 10 documents and a key point (``insight"), the model must correctly identify the two documents that support the point and their associated document IDs. The two facts $f_1,f_2$ may span multiple sentences and may be substantially paraphrased versions of the insight. 

\paragraph{Training Configuration} For each task, we fine-tune two short-context LLMs, Llama-3-8B-Instruct \citep{llama3_1} and Mistral-7B-Instruct-v0.1 \citep{Jiang2023Mistral7}. %
Prior work indicates that attention heads are largely responsible for implementing algorithms \citep{Olsson2022IncontextLA} and using information \textit{within the context} \citep{stolfo-etal-2023-mechanistic,Lieberum2023DoesCA} while MLP layers are responsible for parametric knowledge \citep{geva-etal-2021-transformer}. In addition, when adapting to long contexts, attention heads in particular must handle new position embeddings and softmax over more context tokens \citep{veličković2024softmaxforsharpoutofdistribution}. Therefore, we fine-tune attention heads only.\footnote{We find similar conclusions when fine-tuning all Llama-3-8B-Instruct modules; see Appendix~\ref{sec:appendix_full_finetuning}.} %

To extend models from their original 8K pretrained context length to 32K, we follow \cite{gradient-2024-scaling} in calculating new RoPE \citep{rope}, theta values, using $6315088$ for Llama-3-8B-Instruct and $59300$ for Mistral-7B-Instruct-v0.1. We scale the sliding window accordingly for Mistral-7B-Instruct-v0.1 to 16k context.  These are the only adjustments we make to the models, following \citet{fu2024data}. Our hyperparameters and hardware setup can be found in Appendix~\ref{sec:appendix_training_config_data}.

\begin{figure*}[t!]
\centering
\includegraphics[trim={0.5cm 11cm 3cm 3cm},clip,width=\textwidth]{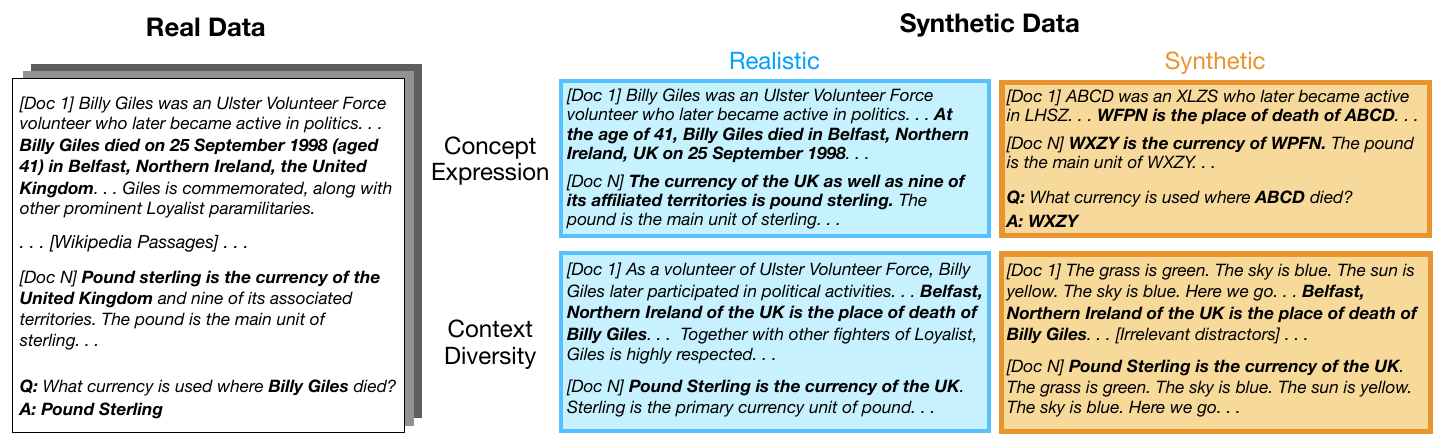}
\caption{Examples of elements of synthetic datasets for MuSiQue with varying levels of \textit{concept expression} and \textit{context diversity}. The needle sentences $f_{i}$ in the context and the entities in them are \textbf{bold}. High concept expression means more realistic expression of the needle $f_{i}$, and low expression means more synthetic, including replacing real entities with symbolic entities or transforming $f_{i}$ into templated sentences. High context diversity means more realistic context surrounding the needles, and low means more synthetic contexts such as repeated, irrelevant padding sentences} %
\label{figure:construction_figure}
\end{figure*}

\section{Synthetic Datasets}

\subsection{Principles Under Consideration}

To create a representative range of synthetic data for each task, we partition the input text $\mathcal{C}$ into (A) text containing relevant information $\{f_1, \ldots, f_m\}$ (``needle \textbf{concepts}") and (B) the surrounding \textbf{context} $C \backslash \{f_1,\ldots,f_m\}$. This allows us to categorize any task as having a variant of \textit{concept expression} (how the target information $f_i$ is expressed) and \textit{context diversity} (the naturalness and relevance of the surrounding information). In the following paragraphs, we discuss common variants found in synthetic data literature. We single out and emphasize a highly structured variant of concept and context, \textit{symbolic tasks}, for being devoid of natural language yet noted to transfer to realistic tasks \mbox{\citep{xiong2024artificial}}. 

\paragraph{Concept Expression} A common procedure for creating synthetic data involves exploiting task asymmetry \citep{josifoski-etal-2023-exploiting, xu-etal-2023-s2ynre, lu-etal-2024-mathgenie, chandradevan-etal-2024-duqgen, chaudhary-etal-2024-relative,Tang-Et-Al:2024:MiniCheck}, where asking an LLM to generate natural language data based off of a label (e.g. a sentence based off of a knowledge triple) is easier than predicting the answer from text of the same complexity and domain. In this scenario, the LLM is asked to create diverse ``needle" target concept expressions $f_i$. In task-specific cases, it is beneficial to make this data less realistic while encouraging generalization. For example, prior synthetic datasets have made use of fictional entities \citep{PrOntoQA} or nonsense phrases \citep{wei2023symbol} in place of real entities and properties, or swapped out nouns to augment the dataset \citep{lu-etal-2024-mathgenie} and prevent overfitting to specific entities. In long context benchmarks \citep{hsieh2024ruler, li2024needlebenchllmsretrievalreasoning}, it is common to express the needle concepts in short, templated sentences.

\paragraph{Context Diversity} We can also vary the expression of $C \backslash \{f_1,\ldots,f_m\}$, the ``haystack.'' This ranges from distractor needles which may have the same form (template) as the target concept to padding with repeated sentences. We use the repeated set of sentences ``The grass is green. The sky is blue. The sun is yellow. Here we go. There and back again." as our low-diversity padding to compare with context that is synthetically generated by an LLM, following \citet{hsieh2024ruler} and \citet{mohtashami2023randomaccess}.

\paragraph{Symbolic Tasks} We also experiment with purely symbolic (involving dictionary key-value or list retrieval) versions of our real tasks, since such tasks are believed to recruit similar model abilities as their natural language counterparts. For example, prior work has indicated that pre-training on code helps on Entity Tracking \citep{prakash2024fine}%
and that fine-tuning on a symbolic dictionary key-value retrieval task can provide greater benefits than even real data \citep{xiong2024artificial}. Additionally, RULER \citep{hsieh2024ruler} introduced a variable assignment task for long-context value tracking. This latter task features expressions like ``\emph{VAR X1 = 12345 ...... VAR Y1 = 54321 ...... Find all variables that are assigned the value 12345.}'' that do not contain meaningful natural language, hence why we differentiate this category from natural language synthetic data.

\begin{table*}[t!]
\caption{Performance (F1) of fine-tuning LLMs on different synthetic data for the long-context retrieval and reasoning tasks. A moderate to large gap exists between the most performant synthetic context extension strategy (\textbf{bold}) and fine-tuning on real data. While careful construction of synthetic data can help close the gap, there does not exist a task-agnostic general way of constructing synthetic datasets for extending LLMs' context window on long-context retrieval and reasoning tasks. $\dagger$ indicates that a model trained on the \textbf{bold} synthetic dataset in the column outperforms a model trained on the indicated dataset with $p < 0.05$ according to a paired bootstrap test. We see that most gains of $\geq 0.02$ accuracy are statistically significant.}
\label{table:synth_perf}
\vspace{0.5em}
\small
\centering
\begin{tabular}{cc|cccc||cc|cc}
\toprule
 \multirow{2}{*}{Concept Exp.} & \multirow{2}{*}{Context Div.} & \multicolumn{2}{c}{MDQA}& \multicolumn{2}{c||}{MuSiQue} &  \multirow{2}{*}{Concept Exp.} & \multirow{2}{*}{Context Div.}   & \multicolumn{2}{c}{SummHay Cite} \\
 &  & \multicolumn{1}{c}{Llama3} & \multicolumn{1}{c}{Mistral} & \multicolumn{1}{c}{Llama3} & \multicolumn{1}{c||}{Mistral} &  &  & \multicolumn{1}{c}{Llama3} & \multicolumn{1}{c}{Mistral} \\
\midrule
High & High & $0.31^\dagger$ & $0.20^\dagger$ & $0.37^\dagger$ & 0.22 & High & High & $0.70^\dagger$ & $0.28^\dagger$ \\
High & Low & $0.41^\dagger$ & $0.23^\dagger$ & \textbf{0.41} & \textbf{0.23} & High & Low & $0.61^\dagger$ & $0.28^\dagger$ \\
Low & High & \textbf{0.49} & \textbf{0.31} & $0.29^\dagger$ & 0.21 & Simplified & High & \textbf{0.79} & \textbf{0.38} \\
Low & Low & $0.47^\dagger$ & $0.24^\dagger$ & $0.34^\dagger$ & $0.17^\dagger$ & Simplified & Low & $0.65^\dagger$ & $0.28^\dagger$\\
Symbolic & Symbolic & 0.48 & $0.16^\dagger$ & $0.32^\dagger$ & $0.11^\dagger$ & Symbolic & Symbolic & $0.54^\dagger$ & $0.18^\dagger$ \\
\midrule
\multicolumn{2}{c|}{Real Data (Full)} &  0.83 & 0.64 & 0.45 & 0.20 & \multicolumn{2}{c|}{\multirow{2}{*}{Real Data (Full)}}  &  \multirow{2}{*}{0.81} &  \multirow{2}{*}{0.40} \\
\multicolumn{2}{c|}{Real Data (Limited)} &  0.80 & 0.59 & 0.32 & 0.16 & & & \multicolumn{1}{c}{} & \multicolumn{1}{c}{} \\

\multicolumn{2}{c|}{Non-FT} &  0.45 & 0.12 & 0.22 & 0.03 & \multicolumn{2}{c|}{Non-FT}  & 0.40 & 0.07 \\
\bottomrule
\end{tabular}
\end{table*}
\subsection{Synthetic Dataset Construction}
\label{sec:synthetic_dataset_construction}

For each of the long-context tasks, we sample a set of examples $\mathcal{D}_{\mathcal{T}}$ from the training set and use the principles above to construct various synthetic datasets based on $\mathcal{D}_{\mathcal{T}}$. See Appendix \ref{sec:appendix_synthetic_data_prompts} for the complete set of prompts used to create the synthetic data, and Appendix \ref{sec:appendix_training_prompts} for our training prompts.

\paragraph{MDQA} Given a training example of MDQA training data $(C,q,y) \in \mathcal{D}_{\mathcal{T}}$, we combine the query $q$ and answer $y$ into our needle $f$ that will be put into the context and that needs to be retrieved by the model. For $f$, we use two simplification levels of concept expression by (1) keeping the real entities in the query and answer (\textbf{high} expression), and (2) replacing the real entities with 4-character symbolic entities (\textbf{low} expression). We create the context surrounding the needle claim with two levels of context diversity: (1) prompting GPT-4o-mini to paraphrase the original context from MDQA training data (for the real entities), or generate a Wikipedia-style paragraph that elaborates on the claim (\textbf{high} diversity); (2) padding the context paragraph with repeated sentences (\textbf{low} diversity). The \textbf{symbolic} dataset is the simple dictionary key-value retrieval dataset from \cite{xiong2024artificial}.

\paragraph{MuSiQue} The $f_i$ here are based on multi-hop knowledge graph relations. Like with MDQA, we create two simplification levels of concept expression by (1) keeping the real entities in the query and answer (\textbf{high} expression), and (2) replacing the real entities with 4-character symbolic entities (\textbf{low} expression), and constructing $f_i$ by prompting GPT-4o-mini to write sentences or via template. %
We create two levels of context diversity by (1) prompting GPT-4 to write a paragraph containing the fact (\textbf{high} diversity), and (2) padding each paragraph with repeated text (\textbf{low} diversity). The \textbf{symbolic} task, as demonstrated in Figure~\ref{fig:symbolic_data}, consists of a list of dictionaries with 4-character identifier, keys and values. Queries are of the form ``What is the \textsc{property\_3} of the \textsc{property\_2} of the \textsc{property\_1} of \textsc{dictionary\_1}?". The answer is found by multi-hop traversal by accessing subsequent dictionary names associated with the specified properties.

\paragraph{SummHay Citation} We derive the $f_i$ from the insights in one of two ways. (1) We prompt GPT-4o-mini to rephrase the insights to create the query, and then prompt again to split rephrased insights into multiple sentences to place into the context (yielding multiple $f_i$ per insight) (\textbf{high} expression); and (2) We prompt GPT-4o-mini to simplify the insights to create the query, and split each simplified insight into multiple sentences to place into the context (\textbf{low} expression). We create two levels of context diversity by (1) padding each document with distractor insights from the same topic, (\textbf{high} diversity) and (2) padding each document with repeated text (\textbf{low} diversity). The \textbf{symbolic} task, as demonstrated in Figure ~\ref{fig:symbolic_data}, consists of lists containing 180 random 4-character strings, where the query is a 4-character string that appears in two different lists. 

\subsection{Results}
Table~\ref{table:synth_perf} shows the performance (F1 scores) of fine-tuning LLMs on different synthetic datasets on the given long-context tasks. We first note that across datasets, \textbf{fine-tuning on synthetic datasets still falls short compared with fine-tuning on real data}, indicating the complexity of the evaluated long-context tasks.\footnote{Particularly on MDQA, we note that such observation is very different from the one in \cite{xiong2024artificial} that finds fine-tuning synthetic data to be more effective than real data. We note that the results of \cite{xiong2024artificial} are obtained on 4K context rather than 32K and the models are fine-tuned with fewer training examples.} For instance, on MuSiQue and SummHay there is a 2-4\% gap between the best synthetic data and real data on Llama 3, and on MDQA there is a much larger gap at 33\%.

Careful construction of synthetic data can help close a lot of the gap by varying the level of concept expression and context diversity beyond the symbolic synthetic dataset. \textbf{However, the effective way of constructing synthetic data for training is very task-specific and can even be counter-intuitive}: there does not exist a single construction strategy that achieves the best performance across tasks, and sometimes a more ``realistic'' synthetic dataset can even underperform the more ``synthetic'' counterparts. 

These results show a complex picture of fine-tuning LLMs with synthetic data for long-context tasks: the downstream performance cannot be simply ``predicted'' by how the synthetic training dataset is constructed. To interpret the success and failure of synthetic data for training, a more fine-grained explanation is needed beyond some general, task-agnostic data construction desiderata. %

\section{Retrieval Heads are Necessary for Context Extension}
\label{sec:retrieval_heads_necessary}

\begin{figure*}[t!]
\begin{center}
\includegraphics[trim={0.3cm 0 0.5cm 0},clip,width=\textwidth]{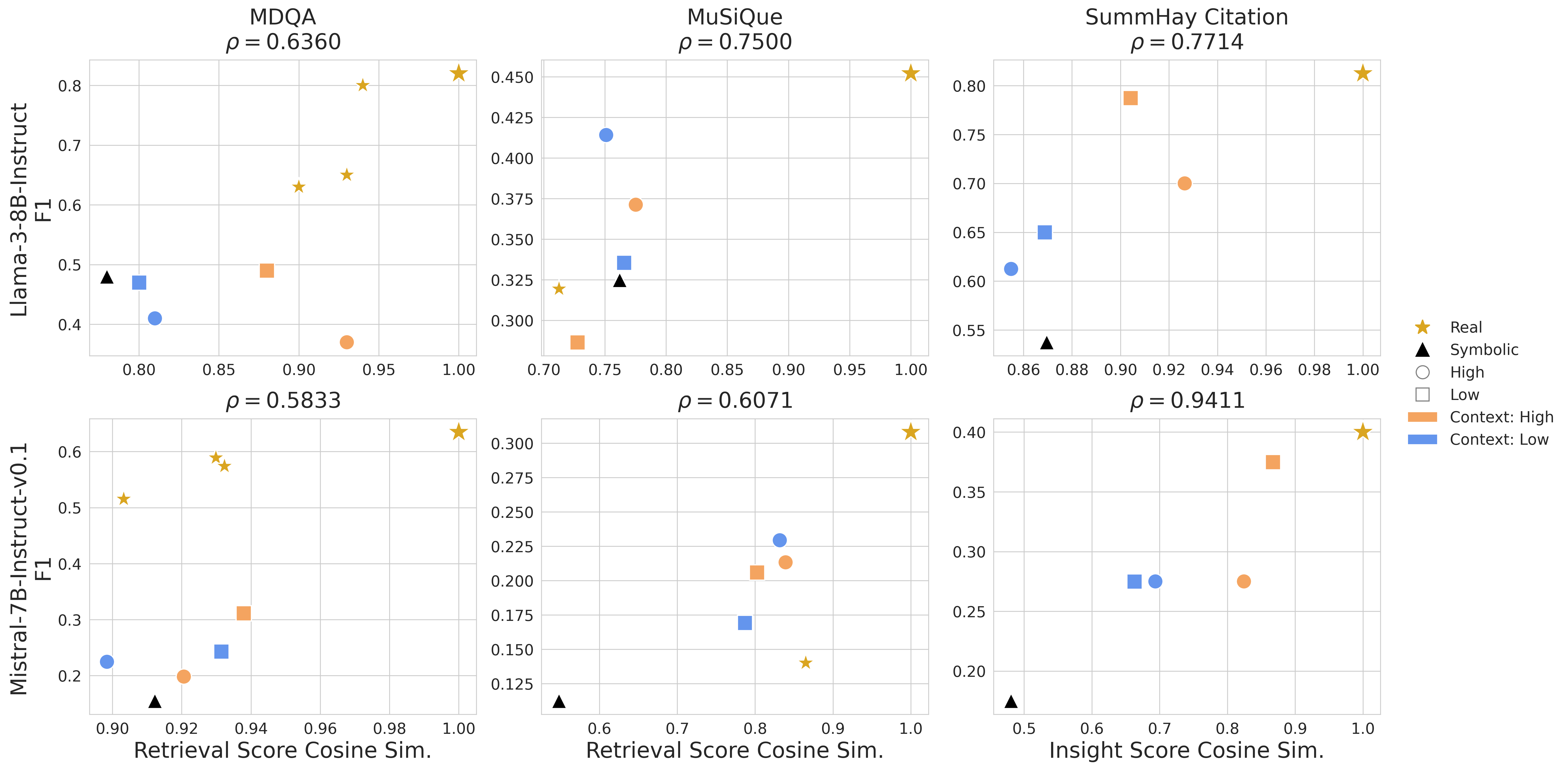}
\end{center}
\caption{Cosine similarity between the retrieval scores on real datasets (R, R) vs. their synthetic versions, and Spearman correlation for each setting. We use multiple limited-relation datasets for MDQA, as described in \mbox{Appendix~\ref{sec:appendix_training_config}}.}
\label{fig:cosine_similarity}
\end{figure*}

One of the key features of our tasks is the need for retrieving needles $f_i$ embedded in a long context. Work from the mechanistic interpretability literature has shown that some attention heads in pre-trained \citep{Olsson2022IncontextLA,Lieberum2023DoesCA} or fine-tuned \citep{tasklocalization,lofit} transformers specialize in retrieving and synthesizing information from the context in principled ways.\footnote{For example, \cite{prakash2024fine} identifies a sparse set of heads that are responsible for retrieving and transmitting the positional information of objects from the context in the entity tracking task.} Notably, recent work \citep{wu2025retrieval} indicates that there exists a special, intrinsic set of attention heads in pre-trained transformers that attend to relevant information $f_i$ in long context $\mathcal{C}$ given a query $q$ and copy it to the output $\mathbf{\tilde{y}}$. \citet{wu2025retrieval} dub them as \textit{retrieval heads}. %

Given the nature of our task, we analyze these attention heads as a proxy for the subnetworks being recruited and learned during fine-tuning with synthetic data. Our core hypothesis is that we can attribute the performance of synthetic context extension to how well the models learn to adapt the attention heads relevant to retrieving and using information from long context, as indicated by the \textit{retrieval scores} of attention heads. Building on prior work, we extend identification of retrieval heads to multi-hop settings in MuSiQue and SummHay Citation. %

\begin{table*}[t!]
\caption{Cosine similarity of real dataset retrieval scores (+ SummHay insight scores) across tasks.}
\vspace{0.5em}
\label{tab:llama3_task_cosine}
\begin{center}

\begin{tabular}{l||cc|cc|cc|cc}
\toprule
 & \multicolumn{2}{c}{MDQA} & \multicolumn{2}{c}{MuSiQue} & \multicolumn{2}{c}{SummHay Retrieval} & \multicolumn{2}{c}{SummHay Insight} \\
 & Llama3 & Mistral & Llama 3 & Mistral & Llama3 & Mistral & Llama3 & Mistral \\
\midrule
MDQA & 1.00 & 1.00 & 0.84 & 0.87 & 0.44 & 0.74 & 0.15 & 0.26 \\
MuSiQue & 0.84 & 0.87 & 1.00 & 1.00 & 0.59 &0.69 & 0.28 & 0.20 \\
SummHay Retrieval & 0.44 & 0.74 & 0.59 &0.69  & 1.00 & 1.00 & 0.08 & 0.07 \\
SummHay Insight & 0.15 & 0.26 & 0.28 & 0.20  & 0.08 &0.07  & 1.00 & 1.00\\
\bottomrule
\end{tabular}

\end{center}
\end{table*}

\subsection{Detecting Retrieval Heads}

Following  \citet{wu2025retrieval}, we detect retrieval heads by computing \emph{retrieval scores}. To compare across fine-tuned models, we consider any attention head with a positive retrieval score to be a \textit{retrieval head}, and later compute cosine similarity to account for the strength of scores. Given a fine-tuned model $\mathcal{M}^{'}$, we evaluate it on a dataset $\mathcal{D}^{*} = \{(\mathcal{C}^{*},q^{*},y^{*})\}$ where the answer $y^{*}$ needs to be identified from some needles $f^{*}$ in $\mathcal{C}^{*}$ and copied to the model output $\tilde{y}$. When $\mathcal{M}^{'}$ generates an output token $w \in \tilde{y}$, we examine whether or not an attention head places the most attention probability mass on the same token in the answer span $y^{*}$ in the context. If so, we consider the token $w$ to be \textit{retrieved} by the attention head. Given an evaluation example $(\mathcal{C}^{*},q^{*},y^{*})$, let $G_{h} = \{w_{h}\}$ be the set of all tokens $w$ that are \textit{retrieved} by a head $h$ during decoding. We define the retrieval score $S_{h}$ for head $h$ on a single example as:
\begin{equation}
    S_{h} = \frac{|G_{h} \cap y^{*}|}{|y^{*}|}
\end{equation}

Note that in the SummHay Citation task, the model is prompted to identify the numerical IDs of the documents (e.g. ``[3]'') that contain the given query insight $q$. In this case, we find it more useful to look at the attention heads that retrieve tokens from the insight needles $f^{*}$ that contain information relevant to $q$ rather than retrieving tokens from the answer $y^{*}$. Note that there are far more tokens in the correct insight needles $f^{*}$ than in the answer $y^{*}$ here. Thus, the \textbf{insight} score for a single example is is defined as:
\begin{equation}
    S_{h} = \mathbbm{1}\left[|G_{h} \cap f^{*}| > 0\right]
\end{equation}

For each head, we average scores over all evaluation examples from $D^{*}$ to yield the final score. 

Given a long-context task $\mathcal{T}$, we detect a set of retrieval heads $H_{\mathrm{real}}$ of the models fine-tuned with real data $\mathcal{D}_{\mathcal{T}}$ on an evaluation set of \textit{real} data . For each model $\mathcal{M}^{'}$ fine-tuned with synthetic data $\tilde{\mathcal{D}}_{\mathcal{T}}$, we detect a set of retrieval heads $H_{\mathrm{synth}}$ on an evaluation set of the corresponding \textit{synthetic} data. $H_{\mathrm{synth}}$ reflects how synthetic context extension enables models to learn modules specialized in retrieving information from \textit{synthetic} long-context data. Following \citet{wu2025retrieval}, we confirm that masking retrieval heads leads to an outsized drop in performance (Figure~\ref{fig:attention_head_masking} in Appendix~\ref{appendix:retrieval_head_masking}). The following sections will examine how $H_{\mathrm{synth}}$ explains synthetic data transferability to \textit{real} long-context data.

\subsection{Case Study} We start with a case study of training Llama-3-8B-Instruct on synthetic data for MuSiQue in Figure~\ref{figure:title_figure}. Highlights show the retrieval score for each head at each layer. The model trained on the real data achieves an F1 score of 0.45 on the evaluation set, and has 129 attention heads which receive a positive retrieval score. Notably, the models trained on synthetic data (both realistic and symbolic) achieve lower F1 (0.41 and 0.33 respectively) while exhibiting far fewer retrieval-scoring attention heads (112 and 74 heads respectively). %
The real data retrieval heads have high recall (0.76 and 0.82) against the synthetic data heads indicating when the synthetic data induces fewer retrieval heads, they tend to be subsets of the real attention heads (\mbox{Appendix~\ref{sec:appendix_recall}}, \mbox{Table~\ref{tab:llama3_musique_recall}}), although this relationship is weaker on MDQA and SummHay Citation.\footnote{We present full retrieval head counts and pairwise recall results in \mbox{Appendix~\ref{sec:appendix_recall}}}

\subsection{Real Task Performance and Retrieval Heads}

As noted previously, when the synthetic data induces fewer retrieval heads, they tend to be from the same population of retrieval heads active on the real data. Following this for each synthetic dataset, we calculate the \textit{recall} of non-zero scoring attention heads against the real dataset (first column of \mbox{Tables~\ref{tab:llama3_mdqa_recall}-\ref{tab:mistral_summhay_insight_recall}} in \mbox{Appendix~\ref{sec:appendix_recall}}). As shown in \mbox{Table~\ref{tab:recall_f1_correlation}}, we find that this is strongly correlated with F1 on the real task for MuSiQue and SummHay Citation. This holds more strongly for Llama-3-8B-Instruct than for Mistral-7B-Instruct-v0.1.

To account for score magnitude, we examine the relationship between the cosine similarity of vectorized retrieval scores with downstream task performance.\footnote{We find it effective to directly match attention heads by index even when models are fine-tuned on different datasets. Visualization in \mbox{Appendix~\ref{sec:appendix_retrieval_hm}} supports this.} Figure~\ref{fig:cosine_similarity} shows a surprisingly strong relationship here. When synthetic data does not induce retrieval heads matching the real task, performance is low. %

This, retrieval heads are important indicators for whether a synthetic dataset targets the desired real data reasoning ability. We view this result as a demonstration of how better understanding of the mechanisms for a target task can allow us to design synthetic tasks that induce the same behavior as required in the real task. However, targeting similar components is not enough; at the same level of similarity, we still observe a wide range of performances, as we will discuss in Section~\ref{sec:synthetic_data_less_effective}.

\subsection{Retrieval Heads Across Tasks}

We ask whether all tasks are leveraging the same set of retrieval heads. Table~\ref{tab:llama3_task_cosine} shows cosine similarity of linearized retrieval scores between tasks. The single-hop and and multi-hop extractive QA tasks, MDQA and MuSiQue, have the highest cosine similarity (Llama-3-8B-Instruct: 0.84; Mistral-7B-Instruct-v0.1: 0.87). However, there is much lower similarity between the QA tasks and the SummHay Citation Retrieval Heads, and the \textit{least} similarity with SummHay Insight Heads.\footnote{SummHay Retrieval Heads attend to the final answer (document number), whereas SummHay Insight Heads attend to the insight text within the document.} %
Comparing to Figure \ref{fig:cosine_similarity}, we find that our real tasks have relatively high cosine similarity ($> 0.66$) with their synthetic versions, with the exception of the purely symbolic chained-dictionary-lookup and list-citation tasks. This suggests that there are task-specific subsets of retrieval heads, either activated based on reasoning ability or token diversity; we leave this for future investigation.

\subsection{Synthetic Data Affects Required Model Components Less Effectively}
\label{sec:synthetic_data_less_effective}
\begin{table}[t]
\caption{Results on Llama-3-8B-Instruct after patching retrieval heads that comprise the complement and intersection between the real and synthetic data versions, compared to random retrieval heads and original performance. Best patching F1 is \underline{underlined}, and best F1 in row is \textbf{bolded}. $\dagger$ indicates that the patched model outperforms the Orig. performance with $p < 0.05$ according to a paired bootstrap test. Patching the intersection outperforms both random and complement heads.}%
\label{tab:llama3_patching}
\begin{center}
\scriptsize

\setlength{\tabcolsep}{5pt}
\begin{tabular}{llllrrrr}
\toprule
\multirow{2}{*}{Task} & \multicolumn{2}{c}{Data Variant} & \multirow{2}{*}{N}  & \multirow{2}{*}{Compl.} & \multirow{2}{*}{Inter.} & \multirow{2}{*}{Rand.} & \multirow{2}{*}{Orig.} \\
 & Concept & Context &  &  &  &  &  \\
\midrule
\midrule
\multirow{6}{*}{MDQA} & Real & Real & - & - & - & - & 0.82 \\
 & Limited & Real & 68 & 0.82 & \textbf{\underline { 0.83}} & 0.82 & 0.80 \\
 & Low & High & 61 & $0.65^\dagger$ & \underline {\boldmath$0.66^\dagger$} & $0.54$ & 0.49 \\
 & Symb. & Symb. & 71 & 0.43 & \underline{\boldmath$0.73^\dagger$} & $0.50^\dagger$ & 0.48 \\
 & Low & Low & 74 & $0.70^\dagger$ & \underline{\boldmath$0.71^\dagger$} & $0.53$ & 0.47 \\
 & High & Low & 74 & $0.63^\dagger$ & $0.51^\dagger$ & \underline{\boldmath$0.70^\dagger$} & 0.41 \\
 & High & High & 60 & $0.52^\dagger$ & \underline {\boldmath$0.59^\dagger$} & 0.26 & 0.37 \\
\midrule
\multirow{6}{*}{MuSiQue} & Real & Real & - & - & - & - & 0.45 \\
 & High & Low & 71 & 0.38 & \underline{\boldmath$0.41^\dagger$} & 0.33 & 0.41 \\
 & High & High & 71 & \underline {0.33} & 0.29 & 0.25 & \textbf{0.37} \\
 & Low & Low & 61 & \underline{\boldmath$0.41^\dagger$} & 0.33 & 0.33 & 0.34 \\
 & Symb. & Symb. & 55 & $0.33^\dagger$ & \underline {\boldmath$0.36^\dagger$} & $0.35^\dagger$ & 0.32 \\
 & Limited & Real & 53 & 0.34 & \underline {\boldmath$0.34$} & 0.31 & 0.32 \\
 & Low & High & 58 & \underline{\boldmath$0.37^\dagger$} & 0.19 & 0.27 & 0.29\\
\midrule
\multirow{6}{*}{SummHay} & Real & Real & - & - & - & - & 0.81 \\
 & Simpl. & High & 27 & 0.73 & \underline {0.74} & 0.73 & \textbf{0.79} \\
 & High & High & 19 & $0.72$ & $0.75$ & \underline{\boldmath$0.79$} & 0.70 \\
 & Simpl. & Low & 26 & $0.66$ & \underline{\boldmath$0.70$} & 0.62 & 0.65 \\
 & High & Low & 21 & $0.57$ & \underline{\boldmath$0.68$} & 0.67 & 0.61 \\
 & Symb. & Symb. & 26 & 0.53 & \textbf{\underline {0.60}} & 0.48 & 0.54 \\
\bottomrule
\end{tabular}

\end{center}
\end{table}

Given different datasets of the same conceptual reasoning and retrieval task, it is peculiar that fine-tuning on some datasets results in fewer retrieval heads, and that synthetic datasets can target subsets of the retrieval heads used for real data. Do the retrieval heads common to all datasets better capture the core capability required for the task? For the common attention heads, do models learn a better way of updating them from the real data than the synthetic data? To investigate these, we follow \citet{prakash2024fine} to perform cross-model activation patching of retrieval heads in the \textit{intersection} and \textit{complement} between the real dataset and the synthetic datasets. Specifically, given the set of retrieval scoring attention heads on the real data, $H_\text{real}$, and the set of retrieval scoring heads on a synthetic dataset, $H_\text{synth}$, we take the complement $H_\text{compl} = H_\text{real}\setminus H_\text{synth}$ and the intersection $H_{\text{inter}} = H_\text{real} \cap H_\text{synth}$. For a fair comparison, we sample $n_\mathrm{heads} = \min(|H_\text{compl}|, |H_\text{inter}|)$ without replacement from both sets. Additionally we compare with $n_\text{heads}$ randomly sampled attention heads. For each set, we patch activations from the model trained on the real data to the model trained on the synthetic data. Implementation details can be found in \mbox{Appendix~\ref{appendix:retrieval_head_patching_details}}.

Our results in Tables \ref{tab:llama3_patching} and \ref{tab:mistral_patching} show that patching \textit{intersection} heads outperforms patching both random and complement heads. The improvement is the greatest for synthetic tasks with the lowest performance on the corresponding real task, and negligible or negative for the best synthetic tasks. The efficacy of patching $H_\text{inter}$ indicates that while a synthetic dataset may target the necessary retrieval heads for the real task, they are \textit{insufficient} in learning how to best utilize the required model components. One explanation is that fine-tuning induces upstream changes so that a different representation distribution is passed to the retrieval heads when learning on synthetic data. This allows retrieval heads to learn to be effective for the synthetic task while failing on out-of-distribution real data representations. %

\section{Related Work}

Prior work has shown that benchmarking or training LLMs on synthetic data can reveal or obtain capabilities that can be transferred and generalized to real tasks, especially in settings where human-annotated data is hard to obtain such as long-context tasks. For this purpose, synthetic data are commonly used and believed to represent a simple reduction of the kinds of abilities employed in linguistically complex settings. The Needle-In-A-Haystack (NIAH) introduced by \cite{kamradt-2023-needle} involves placing a \textit{needle} statement at a random position within a \textit{haystack} consisting of unrelated essay text. Subsequent work \citep{hsieh2024ruler, li2024needlebenchllmsretrievalreasoning} has expanded this task to multi-value retrieval and used simple templated needle sentences to include distractor needles in the context. \cite{hsieh2024ruler} additionally parameterized its test suite by the diversity of the input context and the target value type. These benchmarks are designed in part to expose shortcomings in a model's ability to utilize information throughout its context, known as the \textit{lost-in-the-middle} effect \citep{liu-etal-2024-lost}. To address these shortcomings, prior work has synthesized more realistic long-context data based off of seed corpora to shore up the abilities of models to use their entire pretraining context \citep{an2024make} and further extend up to 1M context \citep{wang2024bootstrap}.

Leveraging the potential generalizability of synthetic data, a line of work in interpretability literature generates synthetic data to perform controlled experiments to probe the inner workings of LLMs. For example, \cite{kim-schuster-2023-entity} shows that a synthetic version of entity tracking can be used to mechanistically understand how fine-tuning enhances existing capabilities of pre-trained LLMs via mechanistic intervention techniques, and \cite{Kim2024AMI} shows that the transformer circuit responsible for syllogistic reasoning in LLMs can be identified by evaluating on synthetic logical statements. However, there is a lack of understanding of when and how the mechanism discovered from synthetic tasks generalizes to real-world capabilities. 

Our work bridges these directions by providing mechanistic explanations for the transferability of synthetic context extension while motivating the pursuit of better usage of synthetic data to evaluate, enhance, and understand the capabilities of LLMs.

\section{Conclusion}

In this paper, we investigated the relationship between the nature of synthetic data for synthetic context extension and performance on downstream tasks. Different synthetic datasets give widely varying performance, partially because of the different numbers of retrieval heads they induce in a model. We showed that these heads are causally connected to the performance, and that these heads are necessary (but not sufficient) for a strong downstream model. We believe this work paves the way for further mechanistic understanding of long context behavior and the ways in which synthetic data induces new capabilities in language models.

\section*{Acknowledgments}

We thank the members of the TAUR lab for their helpful suggestions and discussions throughout this project. We also thank Xi Ye for providing feedback on the draft. This work was partially supported by NSF CAREER Award IIS-2145280, the NSF AI Institute for Foundations of Machine Learning (IFML), the Sloan Foundation via a Sloan Research Fellowship, and a grant from Open Philanthropy. This research has been supported by computing support on the Vista GPU Cluster through the Center for Generative AI (CGAI) and the Texas Advanced Computing Center (TACC) at the University of Texas at Austin.

\section*{Impact Statement}

This paper presents work whose goal is to enable better training of long-context large language models and better understanding of the role of synthetic data in LLM training. There are many potential impacts of better long-context models and training schemes, but none of which are uniquely enabled by the methods and analysis we present here.

\bibliography{example_paper}
\bibliographystyle{icml2025}

\newpage
\appendix
\onecolumn

\section{Synthetic Dataset Creation Prompts}
\label{sec:appendix_synthetic_data_prompts}
\subsection{MDQA}
Given a training example of MDQA data $(\mathcal{C},q,y) \in \mathcal{D}_{\mathcal{T}}$, we first combine the query $q$ and the answer $y$ into a sentence and prompt GPT-4o-mini to rephrase the sentence with the sentence paraphrasing prompt to make it the needle $f$. Then, for the synthetic dataset with high context diversity, we prompt GPT-4o-mini to generate a Wikipedia-style context paragraph with the context generation prompt.

\begin{prompt}[title={Prompt \thetcbcounter: MDQA Sentence Paraphrasing Prompt}]

\promptsubsection{System Prompt} \\ \prompttext{
You are a helpful AI assistant and you are good at creative writing.
}
\\
\\
\promptsubsection{Prompt} \prompttext{
\\
Rewrite the following sentence to Wikipedia style with additional details: \param{{sentence}}
\\
Make sure that readers can correctly answer the following question by reading your rewritten sentence:
\\
Question: \param{{question}}
\\
Answer: \param{{answer}}
}
\end{prompt}

\begin{prompt}[title={Prompt \thetcbcounter: MDQA Context Generation Prompt}]
\promptsubsection{System Prompt} \\ \prompttext{
You are a helpful AI assistant and you are good at creative writing.
}
\\
\\
\promptsubsection{Prompt} \prompttext{
\\
Please make up a 100-word Wikipedia paragraph for the following fake entities: \param{{entity}}. Invent details about people, places, and work related to each entity, and make sure all details are not related to any real-world entities. Give a short, meaningful title to your generated paragraph. After making up the paragraph, please generate a who/when/where/what/why question that:
\\
(1) is related to the given fake entities;
\\
(2) one can use the paragraph to correctly infer the answer within one or two words;
\\
(3) is not a direct copy of a sentence from the paragraph. Please also include the gold answer to the generated question. 
\\
Please give your response in the format:
\\
Title: [title]
\\
Text: [text]
\\
Question: [question]
\\
Answer:[answer]
}
\end{prompt}

\subsection{MuSiQue}

\begin{prompt}[title={Prompt \thetcbcounter: MDQA Sentence Paraphrasing Prompt}]

\promptsubsection{Prompt} \\ \prompttext{
Please make up a single sentence for each of the following fake entities in the style of a wikipedia article.
\\
\param{{fake_entities}}
\\
Please give your response in the format:
\\
Title: [title]
\\
Text: [text]
\\
}
\end{prompt}

\begin{prompt}[title={Prompt \thetcbcounter: MuSiQue Context Generation Prompt}]

\promptsubsection{Prompt} \\ \prompttext{
Please make up a 5-sentence wikipedia paragraph for the following fake entities. Invent details about people, places, and work related to each entity.
\\
\param{{fake_entities}}
\\
Please give your response in the format:
\\
Title: [title]
\\
Text: [text]
\\
}
\end{prompt}

\subsection{SummHay}

\begin{prompt}[title={Prompt \thetcbcounter: SummHay Query Insight (Concept Expression - High) Prompt}]

\promptsubsection{Prompt} \\ \prompttext{
Please rephrase the sentence: ``\param{{text}}"
}
\end{prompt}

\begin{prompt}[title={Prompt \thetcbcounter: SummHay Query Insight (Concept Expression - Simplified) Prompt}]

\promptsubsection{Prompt} \\ \prompttext{
Please simplify and shorten the following sentence. Remove details: ``\param{{sentence}}"
}
\end{prompt}

\begin{prompt}[title={Prompt \thetcbcounter: SummHay Citation Needle Prompt}]
\promptsubsection{Prompt} \\ \prompttext{
"Please break up the following sentence into multiple sentences: ``\param{{text}}"
}
\end{prompt}

\section{Training Prompts}
\label{sec:appendix_training_prompts}

\begin{prompt}[title={Prompt \thetcbcounter: MDQA and MuSiQue Training Prompt}]
\promptsubsection{Prompt} \\ \prompttext{
The following are given passages.
\\
\param{{context}}
\\
Answer the question based on the given passages. Only give me the answer and do not output any other words.
\\
Question: \param{{question}}
\\
Answer:
}
\end{prompt}

\begin{prompt}[title={Prompt \thetcbcounter: SummHay  Citation Training Prompt}]
\promptsubsection{Prompt} \\ \prompttext{
The following are given documents.
\\
\param{{context}}
\\
For the given statement, identify the documents that contain the information by citing the numbers associated with those documents in brackets. For example, if the information in the statement is only found in Document 3, then respond with "[3]". If the information is contained in both Document 3 and Document 7, then respond with "[3][7]". Only output the answer and do not output any other words.
\\
Statement: \param{{statement}}
\\
Answer:
}
\end{prompt}

\section{Additional Data and Training Details}
\label{sec:appendix_training_config}

\subsection{Data}
\label{sec:appendix_training_config_data}
We use 1400 examples for training MDQA models, 400 examples for MuSiQue models, and 400 examples for SummHay Citation models. Each dataset is partitioned in to a 90/10 train/validation split. We use the validation split to calculate retrieval and insight scores.

\begin{tcolorbox}[colback=black!5!white,colframe=black!75!black,title=MDQA Example]

\textbf{Context:}

\prompttext{Document 1: (Title: Don Quixote (Teno)) portion of Don Quixote and his horse are visible. The horse appears to be charging forward out of the stone with his head raised, mouth open, and hooves kicking. The left foot of the horse is not formed, intentionally, by Teno. In Don Quixote's hand is a lance of steel. Both figures are loosely modeled and the figures and stone rest on a oval base measuring which was cut into three pieces for transport by ship to the United States. An inscription on the sculpture reads: King Juan Carlos I and Queen Sofía presented the sculpture June 3, 1976, on\\
...\\
Document 10: (Title: Rocinante) \textcolor{blue}{Rocinante is Don Quixote's horse} in the novel Don Quixote by Miguel de Cervantes. In many ways, Rocinante is not only Don Quixote's horse, but also his double: like Don Quixote, he is awkward, past his prime, and engaged in a task beyond his capacities.\\
...\\
}

\textbf{Question:}

\prompttext{what is don quixote's horse's name}

\textbf{Answer:}

\prompttext{Rocinante}
\end{tcolorbox}

\begin{tcolorbox}[colback=black!5!white,colframe=black!75!black,title=SummHay Citation Example]
\textbf{Context:}

\prompttext{
...\\
Document [24]: ... Furthermore, \textcolor{blue}{the provision of U.S. dollars by global central banks increased, ensuring adequate liquidity within the international financial system. This measure illustrated the depth of the coordinated efforts among major financial institutions to stave off crises and maintain functional stability.} The reverberations of these actions and their impacts on the markets are still unfolding...\\
...\\
Document [27]: ... Turning our gaze to the realm of global financial oversight, central banks are making coordinated efforts to \textcolor{blue}{prevent a liquidity crunch in the international financial system. Recognizing the importance of maintaining robust liquidity, global central banks have ramped up their provision of U.S. dollars}, showcasing a united front in ensuring financial stability. Central banks in Canada, Britain, Japan, Switzerland, and the eurozone have initiated daily currency swaps to ensure that banks operating within their jurisdictions have the necessary dollars to function smoothly. This strategy is aimed at providing stability and fostering confidence in the global banking system during uncertain economic times...\\
...\\
}

\textbf{Statement:}

\prompttext{Global central banks increased their provision of U.S. dollars to ensure adequate liquidity in the international financial system, demonstrating coordinated efforts to prevent a liquidity crunch.}

\textbf{Answer:}

\prompttext{[24][27]}
\end{tcolorbox}

\paragraph{Real Data (Limited)} For MDQA and MuSiQue, we experiment with only training on a subset of the relations involved in eval question hops. 

On MDQA, we create three variants: $\text{L}_1$ is the subset containing Who, When, and Where questions; $\text{L}_2$ is the subset containing When and Where questions; $\text{L}_3$ is the subset containing only Who questions. These comprise 65.8\%, 31.0\%, and 34.8\% of all questions in the MDQA training set respectively. In \mbox{Table~\ref{table:synth_perf}}, \mbox{Table~\ref{tab:llama3_patching}}, and \mbox{Table~\ref{tab:mistral_patching}}, we only report $\text{L}_1$ results due to space constraints. Fine-tuning Llama-3-8B-Instruct on these datasets results in the following F1-scores for the target dataset: $\text{L}_1 = 0.80$, $\text{L}_2 = 0.63$, $\text{L}_3 = 0.65$. Fine-tuning Mistral-7B-Instruct-v0.1 on these datasets results in the following F1-scores for the target dataset: $\text{L}_1 = 0.59$, $\text{L}_2 = 0.52$, $\text{L}_3 = 0.57$.

On MuSiQue, we use the subset of linear 3-hop questions consisting solely of T-REx component questions \mbox{\citep{elsahar-etal-2018-rex}}, as identified by ``$>>$". 10.8\% of MuSiQue linear 3-hop questions in the training set fit this criteria. Additionally, among all component question hops in the training set, 43.0\% are sourced from T-REX.

\subsection{Symbolic Data Construction}

\begin{figure}
    \centering
    \includegraphics[trim={22cm 9.4cm 22cm 8.7cm},clip,width=0.5\textwidth]{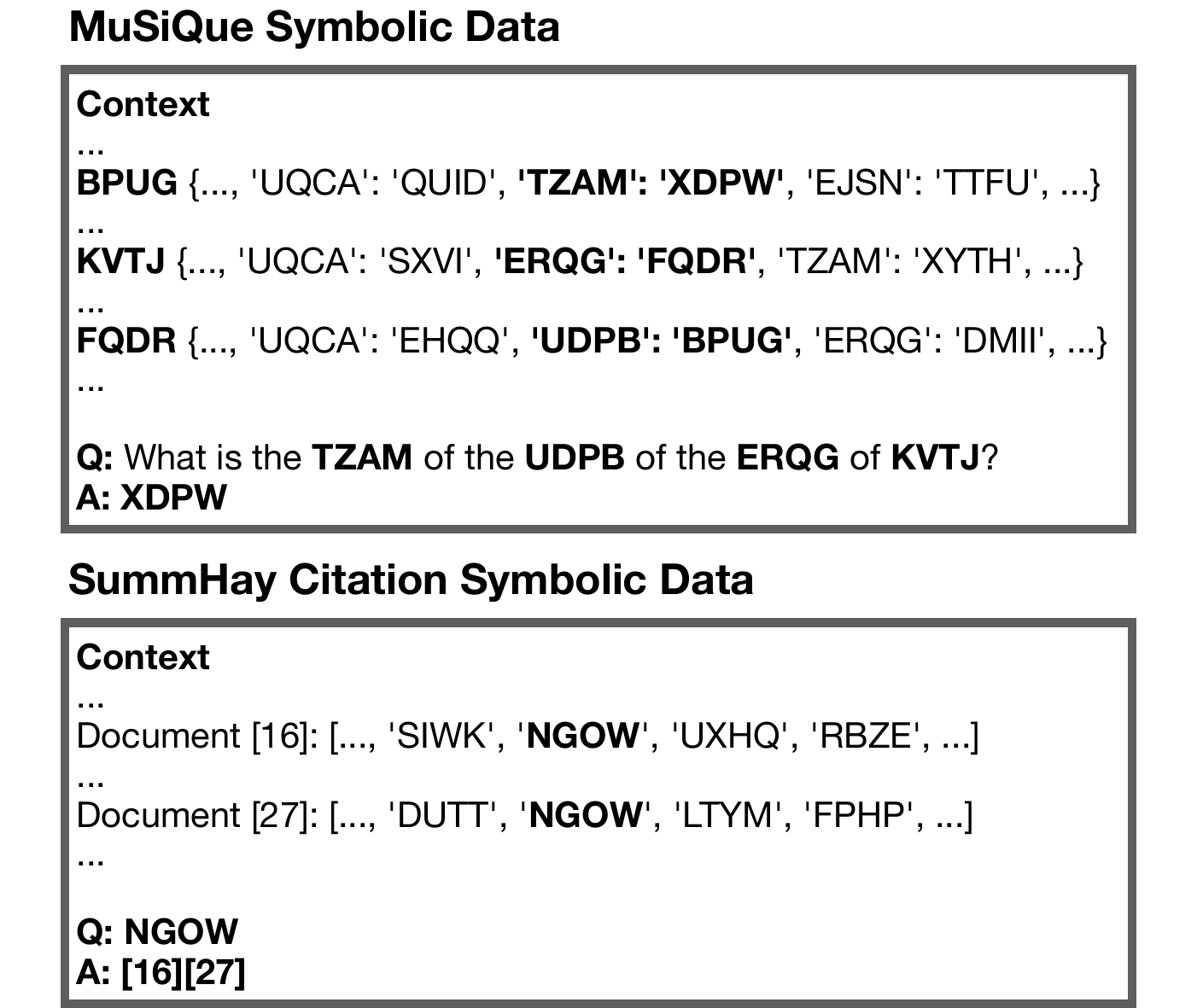}
  \caption{Examples of symbolic data consruction for MuSiQue and SummHay Citation.}
\label{fig:symbolic_data}
\end{figure}

See \mbox{Figure~\ref{fig:symbolic_data}} for examples.

\subsection{Training}
For fine-tuning, we use the Huggingface TRL \citep{von_Werra_TRL_Transformer_Reinforcement} and PEFT \citep{peft} libraries to fine-tune attention heads with LoRA \citep{hu2022lora} (rank = 8 and alpha = 8) using a batch size of 1 and 4 gradient accumulation steps.

We enable Flash Attention 2 and DeepSpeed and use a single NVIDIA H100 GPU (96GB) for each training run. We use greedy decoding in all evaluations.

\section{Retrieval Score Heatmaps}
\label{sec:appendix_retrieval_hm}
\begin{figure}[t!]
\begin{center}
\includegraphics[trim={2.5cm 2.5cm 2.5cm 2.5cm},clip,width=\textwidth]{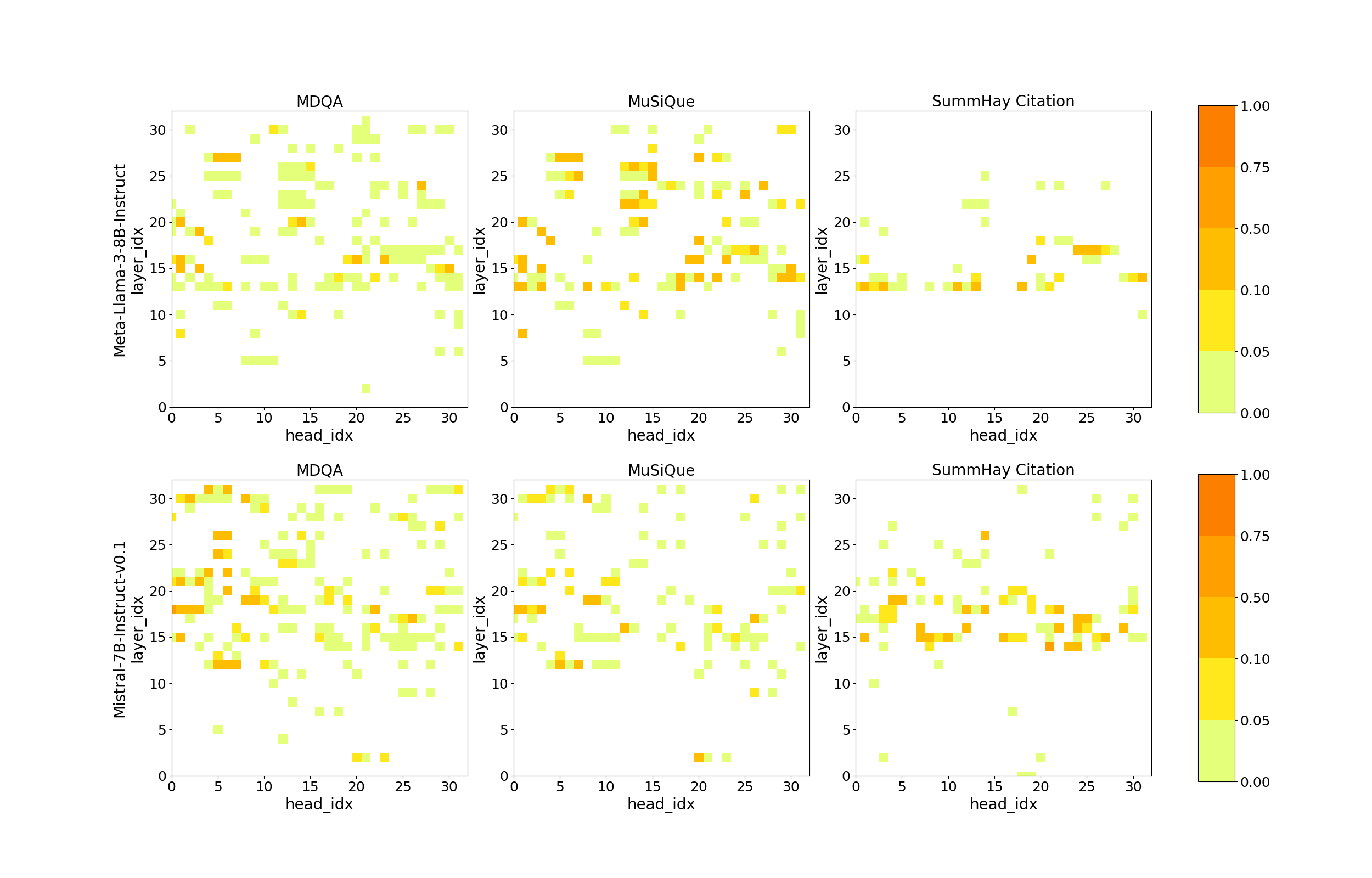}
\end{center}
\caption{Retrieval scores for MDQA, MuSiQue, and Insight scores for SummHay Citation. Top Row: Llama-3-8B-Instruct. Bottom Row: Mistral-7B-Instruct-v0.1. The y-axis indicates the layer index and the x-axis indicates the head index within the layer. We note that retrieval heads are largely found in the last 2/3 layers of the model, as expected according to their involvement in the ``final step" of copying the correct answer to the output. By contrast, SummHay Citation insight heads are concentrated in the middle layers, indicative of their intermediate role. Within a single layer, the specific important attention head indices were likely randomly primed during pretraining to be effectively adapted to the target task.}
\label{fig:task_heatmaps}
\end{figure}

Attention head retrieval scores for the real tasks are shown in Figure \ref{fig:task_heatmaps}.

\begin{figure}[t!]
\begin{center}
\includegraphics[trim={18cm 3cm 16cm 3cm},clip,width=\textwidth]{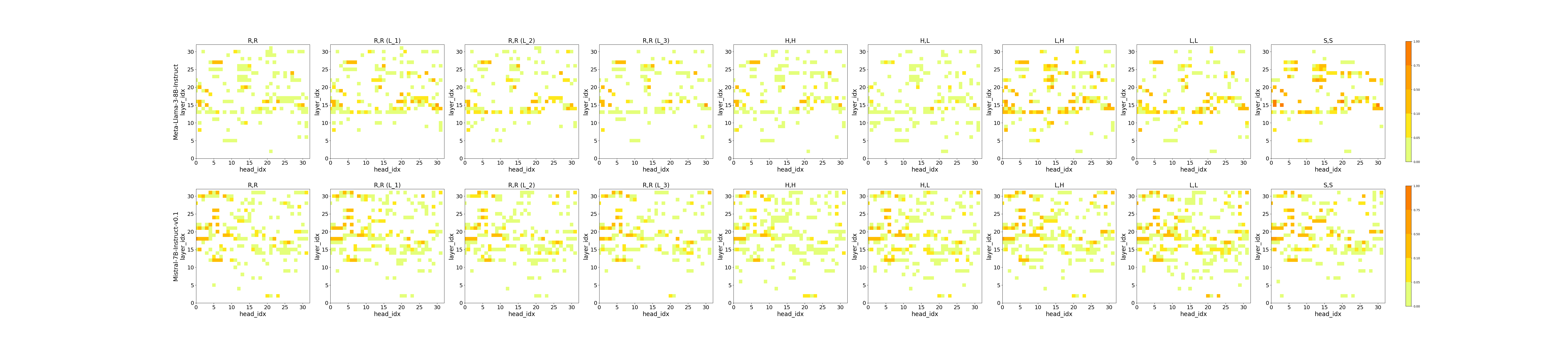}
\end{center}
\caption{Retrieval scores for MDQA and its synthetic dataset versions. Top Row: Llama-3-8B-Instruct. Bottom Row: Mistral-7B-Instruct-v0.1. The y-axis indicates the layer index and the x-axis indicates the head index within the layer.}
\label{fig:variant_mdqa_heatmaps}
\end{figure}

\begin{figure}[t!]
\begin{center}
\includegraphics[trim={15cm 3cm 12cm 3cm},clip,width=\textwidth]{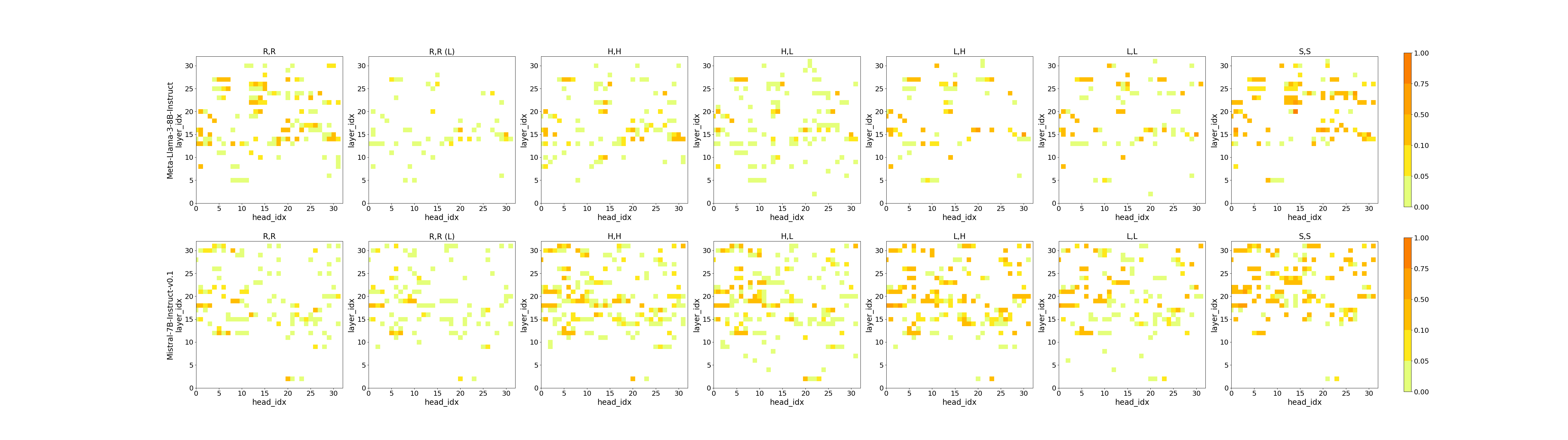}
\end{center}
\caption{Retrieval scores for MuSiQue and its synthetic dataset versions. Top Row: Llama-3-8B-Instruct. Bottom Row: Mistral-7B-Instruct-v0.1. The y-axis indicates the layer index and the x-axis indicates the head index within the layer.}
\label{fig:variant_musique_heatmaps}
\end{figure}

\begin{figure}[t!]
\begin{center}
\includegraphics[trim={12cm 3cm 12cm 3cm},clip,width=\textwidth]{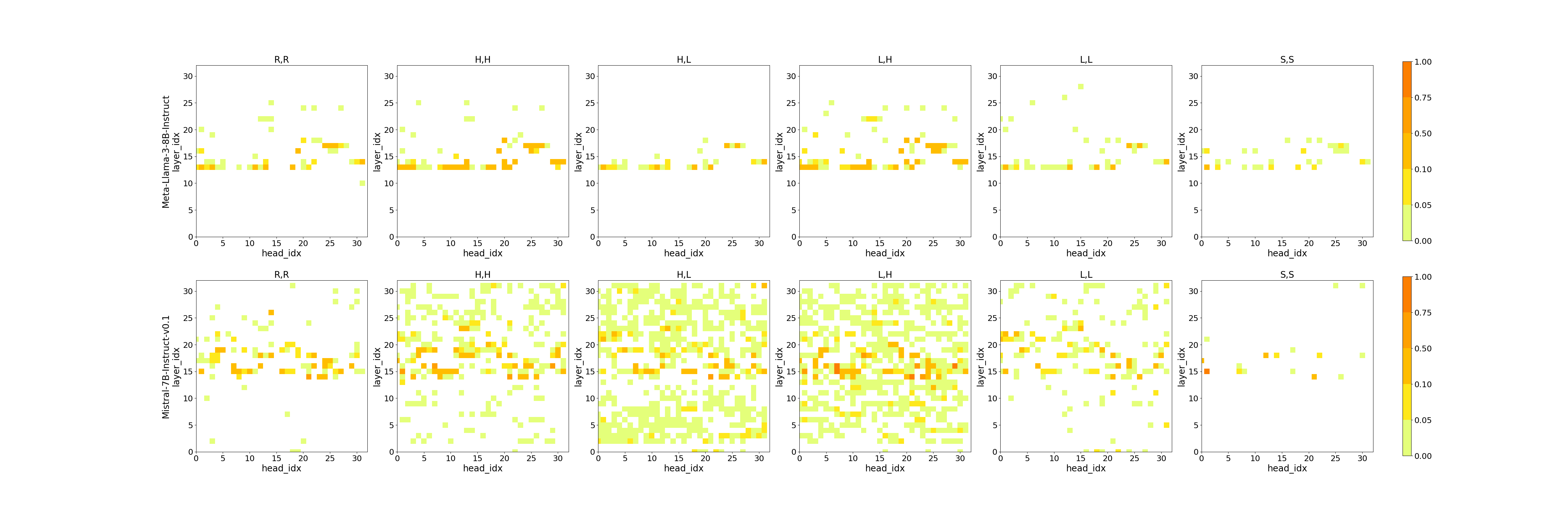}
\end{center}
\caption{Insight scores for SummHay Citation and its synthetic dataset versions. Top Row: Llama-3-8B-Instruct. Bottom Row: Mistral-7B-Instruct-v0.1. The y-axis indicates the layer index and the x-axis indicates the head index within the layer.}
\label{fig:variant_summhay_heatmaps}
\end{figure}

For each target real task, we present heatmaps comparing the real task retrieval scores to the synthetic dataset retrieval scores: MDQA in \mbox{Figure~\ref{fig:variant_mdqa_heatmaps}}, MuSiQue in \mbox{Figure~\ref{fig:variant_musique_heatmaps}}, and SummHay Citation in \mbox{Figure~\ref{fig:variant_summhay_heatmaps}}.

\section{Retrieval Head Recall}
\label{sec:appendix_recall}

In \mbox{Table~\ref{tab:llama3_mdqa_recall}}, \mbox{Table~\ref{tab:llama3_musique_recall}}, and \mbox{Table~\ref{tab:llama3_summhay_insight_recall}}, we examine the overlap between non-zero scoring attention heads on our target tasks and their synthetic versions after fine-tuning Llama-3-8B-Instruct. We find that on all 3 tasks, the attention heads with non-zero retrieval scores on the real data have high recall ($\geq$ 0.76) against those identified on the synthetic data. On MuSiQue and SummHay Citation, we also observe a strong relationship (Spearman R=0.75 and R=1.0 respectively) between the non-zero score attention head recall and F1 on the real task.

However, fine-tuning Mistral-7B-Instruct-v0.1 results in slightly different patterns, as shown in \mbox{Table~\ref{tab:mistral_mdqa_recall}}, \mbox{Table~\ref{tab:mistral_musique_recall}}, and \mbox{Table~\ref{tab:mistral_summhay_insight_recall}}. First, we see more scoring attention heads, which could be caused by the sliding window attention used in the architecture, which only enables a subset of heads to any single position. Second, many of the synthetic datasets result in far more non-zero scoring attention heads, a pattern that we see across all tasks. On MuSiQue and SummHay Citation, we observe a slightly weaker relationship (Spearman R=0.40 and R=0.82 respectively) between the non-zero score attention head recall and F1 on the real task.

\begin{table}[t!]
\begin{center}
\caption{Spearman correlation of synthetic data attention head recall with F1 on the real dataset, showing a strong relationship.}
\label{tab:recall_f1_correlation}
\small
\begin{tabular}{l||rr}
\toprule
 & \multicolumn{2}{c}{Model} \\
 & Llama3 & Mistral \\
\midrule
MDQA & 0.22 & 0.16 \\
MuSiQue & 0.75 & 0.40 \\
SummHay Citation& 1.00 & 0.82 \\
\bottomrule
\end{tabular}
\end{center}
\end{table}

\begin{table}[t!]
\caption{Pairwise recall of Llama-3-8B-Instruct attention heads with non-zero retrieval scores for MDQA synthetic datasets. Limited datasets: $\text{L}_1$ = Who, When, Where; $\text{L}_2$ = When, Where; $\text{L}_3$ = Who. Retrieval Head recall on the real dataset (first column) is weakly correlated with F1 on the real MDQA data (Spearman R = 0.22).}
\label{tab:llama3_mdqa_recall}
\small
\begin{center}

\begin{tabular}{l||rrrrrrrrr|rr}
\toprule
 & \multirow{2}{*}{R,R} & R,R  & R,R & R,R & \multirow{2}{*}{H,H} & \multirow{2}{*}{H,L} & \multirow{2}{*}{L,H} & \multirow{2}{*}{L,L} & \multirow{2}{*}{S,S} & \multirow{2}{*}{\# Heads} & \multirow{2}{*}{F1} \\
 & & ($\text{L}_1$) & ($\text{L}_2$) & ($\text{L}_3 $) & & & & & & &\\
\midrule
R,R & 1.00 & 0.79 & 0.84 & 0.88 & 0.85 & 0.78 & 0.81 & 0.83 & 0.87 & 157 & 0.82 \\
R,R ($\text{L}_1$) & 0.76 & 1.00 & 0.90 & 0.86 & 0.81 & 0.69 & 0.81 & 0.80 & 0.85 & 151 & 0.80 \\
R,R ($\text{L}_2$) & 0.66 & 0.74 & 1.00 & 0.78 & 0.71 & 0.63 & 0.71 & 0.70 & 0.74 & 124 & 0.63 \\
R,R ($\text{L}_3$) & 0.63 & 0.64 & 0.70 & 1.00 & 0.69 & 0.61 & 0.66 & 0.68 & 0.77 & 112 & 0.65 \\
H,H & 0.75 & 0.75 & 0.79 & 0.86 & 1.00 & 0.74 & 0.78 & 0.83 & 0.83 & 139 & 0.37 \\
H,L & 0.73 & 0.68 & 0.74 & 0.80 & 0.78 & 1.00 & 0.77 & 0.80 & 0.78 & 147 & 0.41 \\
L,H & 0.80 & 0.83 & 0.88 & 0.90 & 0.86 & 0.81 & 1.00 & 0.91 & 0.93 & 154 & 0.49 \\
L,L & 0.67 & 0.68 & 0.72 & 0.77 & 0.76 & 0.69 & 0.75 & 1.00 & 0.76 & 127 & 0.47 \\
S,S & 0.64 & 0.66 & 0.69 & 0.79 & 0.69 & 0.61 & 0.70 & 0.69 & 1.00 & 116 & 0.48 \\
\bottomrule
\end{tabular}

\end{center}
\end{table}

\begin{table}[t!]
\caption{Pairwise recall of Llama-3-8B-Instruct attention heads with non-zero retrieval scores for MuSiQue synthetic datasets. We find that the attention heads identified on the real dataset has high recall against all synthetic datasets ($\geq$0.76). Retrieval head recall on the real dataset (first column) is also \textbf{strongly} correlated with F1 on the real MuSiQue data (Spearman R = 0.75).}
\label{tab:llama3_musique_recall}
\small
\begin{center}

\begin{tabular}{l||rrrrrrr|rr}
\toprule
 & R,R & R,R (L) & H,H & H,L & L,H & L,L & S,S & \# Heads & F1 \\
\midrule
R,R & 1.00 & 0.96 & 0.81 & 0.76 & 0.87 & 0.82 & 0.87 & 129 & 0.45 \\
R,R (L) & 0.41 & 1.00 & 0.50 & 0.41 & 0.52 & 0.49 & 0.42 & 55 & 0.32 \\
H,H & 0.59 & 0.85 & 1.00 & 0.63 & 0.70 & 0.65 & 0.63 & 94 & 0.37 \\
H,L & 0.66 & 0.84 & 0.76 & 1.00 & 0.81 & 0.78 & 0.71 & 112 & 0.41 \\
L,H & 0.45 & 0.64 & 0.50 & 0.48 & 1.00 & 0.65 & 0.53 & 67 & 0.29 \\
L,L & 0.47 & 0.65 & 0.51 & 0.52 & 0.72 & 1.00 & 0.55 & 74 & 0.34 \\
S,S & 0.67 & 0.76 & 0.67 & 0.63 & 0.79 & 0.74 & 1.00 & 100 & 0.32 \\
\bottomrule
\end{tabular}

\end{center}
\end{table}

\begin{table}[t!]
\caption{Pairwise recall of Llama-3-8B-Instruct attention heads with non-zero insight scores for SummHay Citation synthetic datasets. Insight head recall on the real dataset (first column) is also \textbf{strongly} correlated with F1 on the real data (Spearman R = 1.0)}
\label{tab:llama3_summhay_insight_recall}
\small
\begin{center}

\begin{tabular}{l||rrrrrr|rr}
\toprule
 & R,R & H,H & H,L & L,H & L,L & S,S & \# Heads & F1 \\
\midrule
R,R & 1.00 & 0.77 & 0.94 & 0.66 & 0.78 & 0.87 & 48 & 0.81 \\
H,H & 0.85 & 1.00 & 1.00 & 0.75 & 0.82 & 0.87 & 53 & 0.70 \\
H,L & 0.60 & 0.58 & 1.00 & 0.48 & 0.72 & 0.70 & 31 & 0.61 \\
L,H & 0.90 & 0.92 & 1.00 & 1.00 & 0.88 & 0.93 & 65 & 0.79 \\
L,L & 0.65 & 0.62 & 0.94 & 0.54 & 1.00 & 0.80 & 40 & 0.65 \\
S,S & 0.54 & 0.49 & 0.68 & 0.43 & 0.60 & 1.00 & 30 & 0.54 \\
\bottomrule
\end{tabular}
\end{center}
\end{table}

\begin{table}[t!]
\caption{Pairwise recall of Mistral-7B-Instruct-v0.1 attention heads with non-zero retrieval scores for MDQA synthetic datasets. Limited datasets: $\text{L}_1$ = Who, When, Where; $\text{L}_2$ = When, Where; $\text{L}_3$ = Who. Retrieval head recall on the real dataset (first column) is weakly correlated with F1 on the real MDQA data (Spearman R = 0.16).}
\label{tab:mistral_mdqa_recall}
\small
\begin{center}

\begin{tabular}{l||rrrrrrrrr|rr}
\toprule
 & \multirow{2}{*}{R,R} & R,R  & R,R & R,R & \multirow{2}{*}{H,H} & \multirow{2}{*}{H,L} & \multirow{2}{*}{L,H} & \multirow{2}{*}{L,L} & \multirow{2}{*}{S,S} & \multirow{2}{*}{\# Heads} & \multirow{2}{*}{F1} \\
 & & ($\text{L}_1$) & ($\text{L}_2$) & ($\text{L}_3 $) & & & & & & &\\
\midrule
R,R & 1.00 & 0.76 & 0.76 & 0.81 & 0.74 & 0.74 & 0.77 & 0.69 & 0.80 & 178 & 0.63 \\
R,R ($\text{L}_1$) & 0.81 & 1.00 & 0.85 & 0.86 & 0.77 & 0.80 & 0.80 & 0.72 & 0.82 & 192 & 0.59 \\
R,R ($\text{L}_2$) & 0.81 & 0.84 & 1.00 & 0.86 & 0.74 & 0.77 & 0.80 & 0.72 & 0.87 & 190 & 0.52 \\
R,R ($\text{L}_3$) & 0.74 & 0.72 & 0.73 & 1.00 & 0.67 & 0.71 & 0.72 & 0.63 & 0.74 & 161 & 0.57 \\
H,H & 0.86 & 0.84 & 0.81 & 0.86 & 1.00 & 0.83 & 0.83 & 0.76 & 0.84 & 208 & 0.20 \\
H,L & 0.88 & 0.89 & 0.86 & 0.93 & 0.85 & 1.00 & 0.86 & 0.81 & 0.87 & 212 & 0.22 \\
L,H & 0.88 & 0.85 & 0.86 & 0.92 & 0.82 & 0.83 & 1.00 & 0.78 & 0.85 & 205 & 0.31 \\
L,L & 0.93 & 0.91 & 0.91 & 0.94 & 0.88 & 0.92 & 0.91 & 1.00 & 0.91 & 240 & 0.24 \\
S,S & 0.80 & 0.76 & 0.81 & 0.81 & 0.72 & 0.73 & 0.74 & 0.68 & 1.00 & 178 & 0.15 \\
\bottomrule
\end{tabular}
\end{center}
\end{table}

\begin{table}[t!]
\caption{Pairwise recall of Mistral-7B-Instruct-v0.1 attention heads with non-zero retrieval scores for MuSiQue synthetic datasets. Retrieval Head recall on the real dataset (first column) is also moderately correlated with F1 on the real MuSiQue data (Spearman R = 0.40)}
\label{tab:mistral_musique_recall}
\small
\begin{center}
\begin{tabular}{l||rrrrrrr|rr}
\toprule
 & R,R & R,R (L) & H,H & H,L & L,H & L,L & S,S & \# Heads & F1 \\
\midrule
R,R & 1.00 & 0.66 & 0.56 & 0.55 & 0.63 & 0.66 & 0.62 & 111 & 0.31 \\
R,R (L) & 0.63 & 1.00 & 0.57 & 0.57 & 0.59 & 0.60 & 0.53 & 106 & 0.14 \\
H,H & 0.89 & 0.95 & 1.00 & 0.83 & 0.88 & 0.86 & 0.82 & 178 & 0.21 \\
H,L & 0.83 & 0.91 & 0.78 & 1.00 & 0.81 & 0.81 & 0.78 & 167 & 0.23 \\
L,H & 0.84 & 0.83 & 0.73 & 0.72 & 1.00 & 0.82 & 0.80 & 148 & 0.21 \\
L,L & 0.75 & 0.72 & 0.61 & 0.61 & 0.70 & 1.00 & 0.68 & 126 & 0.17 \\
S,S & 0.74 & 0.66 & 0.61 & 0.62 & 0.72 & 0.72 & 1.00 & 133 & 0.11 \\
\bottomrule
\end{tabular}
\end{center}
\end{table}

\begin{table}[t!]
\caption{Pairwise recall of Mistral-7B-Instruct-v0.1 attention heads with non-zero insight scores for SummHay Citation synthetic datasets. Insight Head recall on the real dataset (first column) is also \textbf{strongly} correlated with F1 on the real SummHay Citation data (Spearman R = 0.82)}
\label{tab:mistral_summhay_insight_recall}
\small
\begin{center}

\begin{tabular}{l||rrrrrr|rr}
\toprule
 & R,R & H,H & H,L & L,H & L,L & S,S & \# Heads & F1 \\
\midrule
R,R & 1.00 & 0.31 & 0.15 & 0.16 & 0.45 & 0.81 & 91 & 0.40 \\
H,H & 0.87 & 1.00 & 0.34 & 0.39 & 0.75 & 0.88 & 259 & 0.28 \\
H,L & 0.79 & 0.65 & 1.00 & 0.57 & 0.80 & 0.81 & 496 & 0.28 \\
L,H & 0.86 & 0.76 & 0.57 & 1.00 & 0.78 & 0.88 & 497 & 0.38 \\
L,L & 0.75 & 0.44 & 0.24 & 0.24 & 1.00 & 0.81 & 150 & 0.28 \\
S,S & 0.14 & 0.05 & 0.03 & 0.03 & 0.09 & 1.00 & 16 & 0.17 \\
\bottomrule
\end{tabular}
\end{center}
\end{table}

\section{Retrieval Head Masking}
\label{appendix:retrieval_head_masking}

\begin{figure*}[t!]
\begin{center}
\includegraphics[trim={0 0.25cm 0 0.25cm},clip,width=0.95\textwidth]{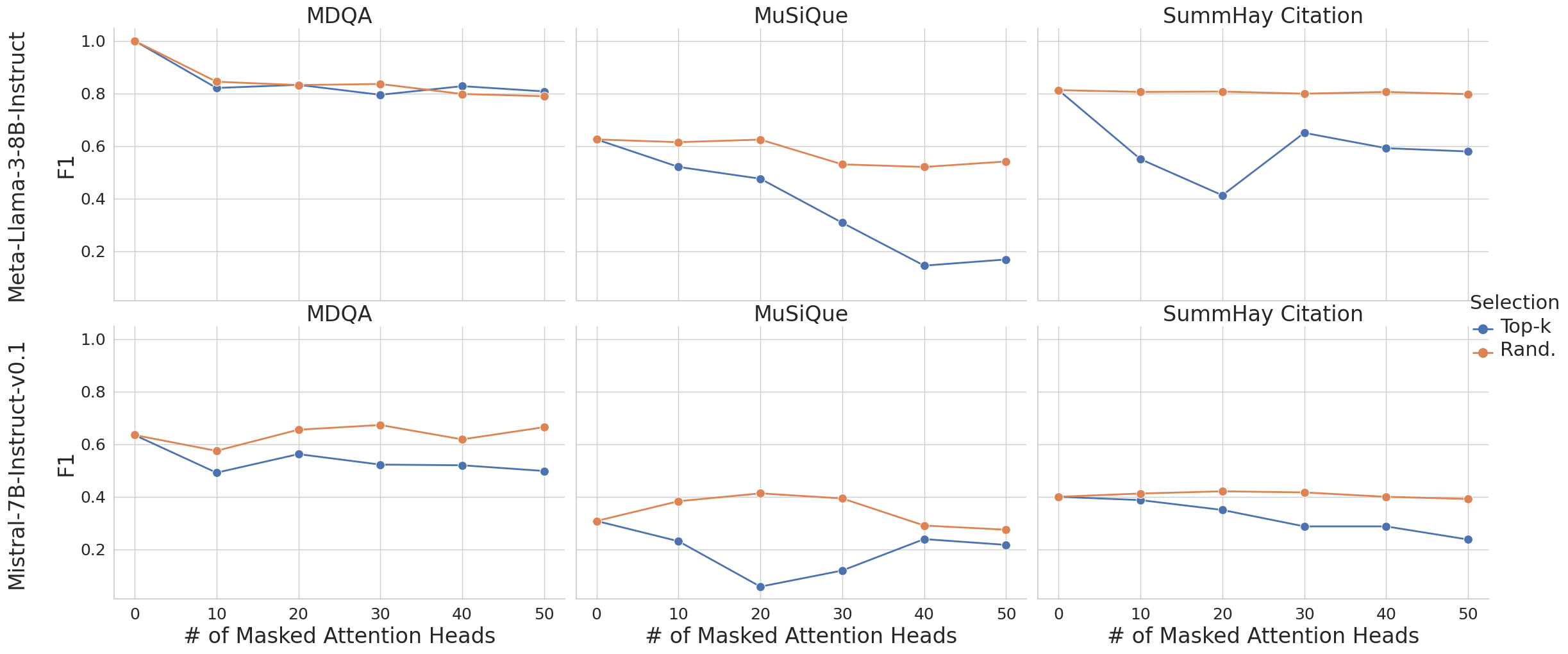}
\end{center}
\caption{Top row: Llama-3-8B-Instruct. Bottom row: Mistral-7B-Instruct-v0.1. Effect of masking activations from attention heads (following \citet{wu2025retrieval}) with the top-$k$ highest retrieval (MDQA, MuSiQue) or insight (SummHay Citation) scores. We compare with masking the same number of randomly chosen heads, averaged over 3 samples. Masking top-$k$ attention heads consistently results in a larger drop in performance than masking random attention heads.}
\label{fig:attention_head_masking}
\end{figure*}

Figure~\ref{fig:attention_head_masking} shows the effect of masking attention heads with the top-$k$ retrieval (MDQA, MuSiQUE) or insight (SummHay Citation) scores. Compared to masking the same number of randomly chosen heads (over 3 trials), masking top-$k$ attention heads consistently results in a larger drop in performance than masking random attention heads.

\section{Retrieval Head Patching Details}
\label{appendix:retrieval_head_patching_details}

We implemented retrieval head patching with Baukit.\footnote{\url{https://github.com/davidbau/baukit}} Given an example from the test set and a set of attention heads to patch, we run a forward pass with the model fine-tuned on the real data and extract the attention output from the selected attention heads before being projected and concatenated back to the residual stream. Then, we use the same example and run a forward pass with the model fine-tuned on a synthetic dataset. We replace the attention outputs of the aforementioned selected attention heads with the attention outputs extracted from the model fine-tuned on real data. Using the procedure described above, we patch the attention outputs of the selected attention heads into the model fine-tuned on a synthetic dataset for \emph{all input} tokens. We then use the patched inputs to generate and decode output tokens without patching any activations for the output tokens.

\subsection{Mistral-7B-Instruct-v0.1 Retrieval Head Patching}
\label{sec:appendix_mistral_patching}

\begin{table}[t]
\caption{Results on Mistral-7B-Instruct-v0.1 after patching heads that comprise the complement and intersection retrieval heads between the real and synthetic data versions, compared to random retrieval heads and original performance. The best patching F1 is \underline{underlined}, and the best F1 in each row is \textbf{bolded}. $\dagger$ indicates that the patched model outperforms the Orig. performance with $p < 0.05$ according to a paired bootstrap test. Patching the intersection outperforms both random and complement heads.}%
\label{tab:mistral_patching}
\begin{center}
\scriptsize

\begin{tabular}{llllrrrr}
\toprule
\multirow{2}{*}{Task} & \multicolumn{2}{c}{Data Variant} & \multirow{2}{*}{N}  & \multirow{2}{*}{Compl.} & \multirow{2}{*}{Inter.} & \multirow{2}{*}{Rand.} & \multirow{2}{*}{Orig.}\\
 & Concept & Context &  &  &  &  & \\
\midrule
\midrule
\multirow{6}{*}{MDQA} & Real & Real & - & - & - & - & 0.63 \\
 & Limited & Real & 80 & \underline {0.57} & 0.53 & 0.49 & \textbf{0.59} \\
 & Low & High & 69 & 0.21 & \underline{\boldmath$0.34^\dagger$} & 0.23 & 0.31 \\
 & Low & Low & 86 & 0.21 & \underline{\boldmath$0.40^\dagger$} & 0.26 & 0.24 \\
 & High & Low & 78 & 0.13 & \underline{\boldmath$0.31^\dagger$} & 0.16 & 0.22 \\
 & High & High & 80 & 0.21 & \underline{\boldmath$0.26^\dagger$} & 0.18 & 0.20 \\
 & Symb. & Symb. & 70 & 0.01 & \underline {0.02} & 0.02 & \textbf{0.15} \\
\midrule
\multirow{6}{*}{MuSiQue} & Real & Real & - & - & - & - & 0.31 \\
 & High & Low & 92 & $0.23^\dagger$ & \underline{\boldmath$0.26^\dagger$} & 0.20 & 0.23 \\
 & High & High & 91 & 0.16 & \underline{\boldmath$0.24^\dagger$} & 0.20 & 0.21 \\
 & Low & High & 73 & 0.14 & \underline{\boldmath$0.21^\dagger$} & 0.17 & \textbf{0.21} \\
 & Low & Low & 71 & 0.15 & \underline{\boldmath$0.18^\dagger$} & 0.16 & 0.17 \\
 & Limited & Real & 70 & 0.14 & \underline{\boldmath$0.20$} & 0.18 & 0.14 \\
 & Symb. & Symb. & 80 & $0.14^\dagger$ & \underline{\boldmath$0.19^\dagger$} & $0.15^\dagger$ & 0.11 \\
\midrule
\multirow{6}{*}{SummHay} & Real & Real & - & - & - & - & 0.40 \\
 & Simpl. & High & 78 &  0.34 & \underline{\boldmath$0.35$} & 0.3 & 0.38 \\
 & High & Low & 72 & 0.33 & 0.33 & \underline{\boldmath$0.35$} & 0.28 \\
 & High & High & 70 & \underline{\boldmath$0.30$}& 0.30 & 0.29 & 0.28 \\
 & Simpl. & Low & 68 & 0.29 & \underline{\boldmath$0.33$} & 0.30 & 0.28 \\
 & Symb. & Symb. & 13 & 0.14 & 0.14 & \underline {0.16} & \textbf{0.18} \\
\bottomrule
\end{tabular}

\end{center}
\end{table}

See Table~\ref{tab:mistral_patching}.

\subsection{Intersection and Complement Head Retrieval Scores}
\begin{table}[t!]
\caption{Average retrieval / insight scores for attention heads in the intersection and the complement.}
\label{tab:inter_compl_retrieval_scores}
\small
\begin{center}

\begin{tabular}{lllrrrr}
\toprule
\multirow{2}{*}{Task}& \multicolumn{2}{c}{Dataset Variant} & \multicolumn{2}{c}{Llama-3-8B-Instruct} & \multicolumn{2}{c}{Mistral-7B-Instruct-v0.1}\\
& Concept & Context & Inter. & Compl. & Inter. &  Compl. \\
\midrule
\multirow{8}{*}{MDQA} 
& Real & Real (Who, When, Where) & 0.045 & 0.011 & 0.059 & 0.019 \\
& Real & Real (Who) & 0.052 & 0.012 & 0.062 & 0.021 \\
& Real & Real (When, Where) & 0.050 & 0.012 & 0.059 & 0.018 \\
& High & High & 0.046 & 0.010 & 0.057 & 0.020 \\
& High & Low & 0.045 & 0.015 & 0.056 & 0.018 \\
& Low & High & 0.044 & 0.010 & 0.057 & 0.013 \\
& Low & Low & 0.049 & 0.013 & 0.054 & 0.015 \\
& Symbolic & Symbolic & 0.051 & 0.013 & 0.059 & 0.020 \\
\midrule
\multirow{6}{*}{MuSiQue}
& Real & Real (Limited) & 0.121 & 0.049 & 0.065 & 0.021 \\
& High & High & 0.105 & 0.040 & 0.053 & 0.015 \\
& High & Low & 0.096 & 0.045 & 0.055 & 0.018 \\
& Low & High & 0.116 & 0.048 & 0.055 & 0.017 \\
& Low & Low & 0.106 & 0.054 & 0.058 & 0.021 \\
& Symbolic & Symbolic & 0.099 & 0.037 & 0.057 & 0.026 \\
\midrule
\multirow{5}{*}{SummHay}
& High & High & 0.071 & 0.008 & 0.093 & 0.036 \\
& High & Low & 0.092 & 0.016 & 0.098 & 0.039 \\
& Simplified & Low & 0.087 & 0.017 & 0.097 & 0.050 \\
& Simplified & High & 0.068 & 0.008 & 0.093 & 0.041 \\
& Symbolic & Symbolic & 0.097 & 0.021 & 0.213 & 0.064 \\
\bottomrule
\end{tabular}
\end{center}
\end{table}

See Table~\ref{tab:inter_compl_retrieval_scores}.

\section{Full Finetuning}
\label{sec:appendix_full_finetuning}

In this section, we present results on Meta-Llama-3-8B-Instruct with fine-tuning of all LoRA modules, and demonstrate that we find similar conclusions.

\subsection{Synthetic Data Performance}

See Table~\ref{table:synth_perf_llama3_fullft}. We find that there are mostly small ($< 0.05$) performance differences between fine-tuning only attention heads and all modules. Notable exceptions are found in the SummHay Citation task, where the performance of the synthetic datasets increase up to +0.13 (High, High).

\begin{table}[t!]
\caption{Llama-3-8B-Instruct (all LoRA modules): Performance (F1) of fine-tuning  on different synthetic data on the long-context retrieval and reasoning tasks. The results of training on the best synthetic datasets are bolded.}
\label{table:synth_perf_llama3_fullft}

\begin{center}
\small
\begin{tabular}{cc|cc||cc|c}
\toprule
 \multirow{1}{*}{Concept Exp.} & \multirow{1}{*}{Context Div.} & \multicolumn{1}{c}{MDQA}& \multicolumn{1}{c||}{MuSiQue} &  \multirow{1}{*}{Concept Exp.} & \multirow{1}{*}{Context Div.}   & \multicolumn{1}{c}{SummHay} \\
\midrule
High & High & 0.35 & 0.40 & High & High & 0.83  \\
High & Low & 0.39 & \textbf{0.42} & High & Low & 0.68 \\
Low & High & \textbf{0.49} & 0.30 & Simplified & High & \textbf{0.83} \\
Low & Low & 0.47 & 0.38 & Simplified & Low & 0.58 \\
Symbolic & Symbolic & 0.46 & 0.37 & Symbolic & Symbolic & 0.63 \\
\midrule
\multicolumn{2}{c|}{Real Data (Full)} &  0.82 & 0.45 & \multicolumn{2}{c|}{\multirow{2}{*}{Real Data (Full)}}  &  \multirow{2}{*}{0.81} \\
\multicolumn{2}{c|}{Real Data (Limited)} &  0.84 & 0.32 & & & \\
\bottomrule
\end{tabular}
\end{center}
\end{table}

\subsection{Retrieval Score Heatmaps}

\begin{figure}[t!]
\begin{center}
\includegraphics[trim={2.5cm 1cm 2.5cm 1cm},clip,width=\textwidth]{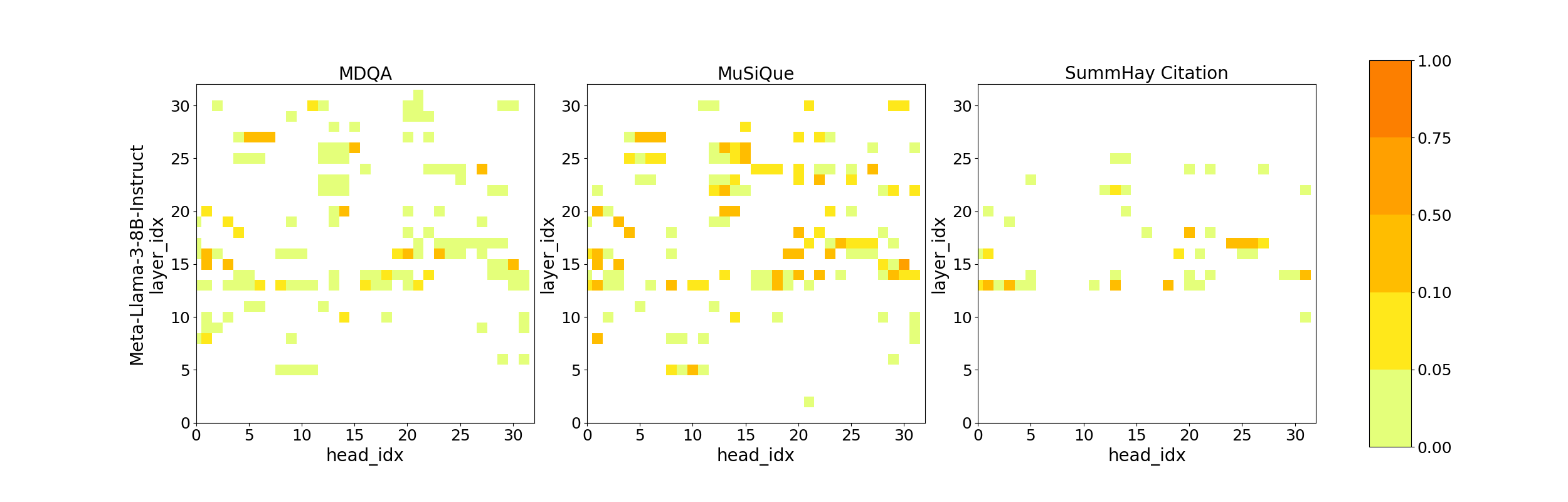}
\end{center}
\caption{Llama-3-8B-Instruct (all LoRA modules): Retrieval scores for MDQA, MuSiQue, and Insight scores for SummHay Citation, after fine-tuning on each task. The y-axis indicates the layer index and the x-axis indicates the head index within the layer.}
\label{fig:task_heatmaps_llama3_fullft}
\end{figure}

See Figure~\ref{fig:task_heatmaps_llama3_fullft}.

\subsection{Retrieval Head Recall}
\label{sec:appendix_llama3_fullft_recall}

In Table~\ref{tab:llama3_fullft_mdqa_recall}, Table~\ref{tab:llama3_fullft_musique_recall}, and Table~\ref{tab:llama3_fullft_summhay_insight_recall}, we find that there are generally fewer non-zero scoring attention heads on the synthetic tasks, compared to the real task. On MuSiQue, the non-zero attention heads tend to be subsets of the those identified on the real task, as when only fine-tuning attention modules.

\begin{table}[t!]
\caption{Llama-3-8B-Instruct (LoRA all modules): Pairwise recall of attention heads with non-zero retrieval scores for MDQA synthetic datasets. Limited datasets: $\text{L}_1$ = Who, When, Where; $\text{L}_2$ = When, Where; $\text{L}_3$ = Who. Recall of real data retrieval heads is moderately correlated with F1 (Spearman R = 0.60).}
\label{tab:llama3_fullft_mdqa_recall}
\small
\begin{center}
\begin{tabular}{l||rrrrrrrrr|rr}
\toprule
 & R,R & R,R ($\text{L}_1$) & R,R ($\text{L}_2$) & R,R ($\text{L}_3$) & H,H & H,L & L,H & L,L & S,S & \# Heads & F1 \\
\midrule
R,R & 1.00 & 0.71 & 0.70 & 0.76 & 0.79 & 0.73 & 0.71 & 0.75 & 0.76 & 137 & 0.82 \\
R,R ($\text{L}_1$) & 0.77 & 1.00 & 0.83 & 0.84 & 0.79 & 0.71 & 0.80 & 0.83 & 0.84 & 148 & 0.84 \\
R,R ($\text{L}_2$) & 0.77 & 0.84 & 1.00 & 0.83 & 0.79 & 0.71 & 0.75 & 0.80 & 0.84 & 150 & 0.73 \\
R,R ($\text{L}_3$) & 0.72 & 0.74 & 0.71 & 1.00 & 0.69 & 0.67 & 0.70 & 0.75 & 0.77 & 129 & 0.72 \\
H,H & 0.73 & 0.67 & 0.66 & 0.67 & 1.00 & 0.69 & 0.70 & 0.74 & 0.74 & 126 & 0.35 \\
H,L & 0.76 & 0.69 & 0.67 & 0.74 & 0.79 & 1.00 & 0.72 & 0.78 & 0.76 & 143 & 0.39 \\
L,H & 0.82 & 0.86 & 0.79 & 0.86 & 0.89 & 0.80 & 1.00 & 0.91 & 0.90 & 159 & 0.49 \\
L,L & 0.76 & 0.77 & 0.74 & 0.81 & 0.81 & 0.76 & 0.79 & 1.00 & 0.82 & 138 & 0.47 \\
S,S & 0.69 & 0.71 & 0.70 & 0.74 & 0.73 & 0.66 & 0.70 & 0.75 & 1.00 & 125 & 0.46 \\
\bottomrule
\end{tabular}
\end{center}
\end{table}

\begin{table}[t!]
\caption{Llama-3-8B-Instruct (LoRA all modules): Pairwise recall of attention heads with non-zero retrieval scores for MuSiQue synthetic datasets. Recall of real data retrieval heads is moderately correlated with F1 (Spearman R = 0.36).}
\label{tab:llama3_fullft_musique_recall}
\small
\begin{center}

\begin{tabular}{l||rrrrrrr|rr}
\toprule
 & R,R & R,R (L) & H,H & H,L & L,H & L,L & S,S & \# Heads & F1 \\
\midrule
R,R & 1.00 & 0.94 & 0.82 & 0.77 & 0.88 & 0.88 & 0.86 & 135 & 0.48 \\
R,R (L) & 0.44 & 1.00 & 0.56 & 0.41 & 0.46 & 0.56 & 0.39 & 63 & 0.41 \\
H,H & 0.59 & 0.87 & 1.00 & 0.64 & 0.75 & 0.73 & 0.61 & 98 & 0.40 \\
H,L & 0.78 & 0.89 & 0.89 & 1.00 & 0.90 & 0.86 & 0.78 & 136 & 0.42 \\
L,H & 0.65 & 0.73 & 0.77 & 0.66 & 1.00 & 0.83 & 0.71 & 100 & 0.30 \\
L,L & 0.50 & 0.68 & 0.57 & 0.49 & 0.64 & 1.00 & 0.55 & 77 & 0.38 \\
S,S & 0.75 & 0.73 & 0.73 & 0.68 & 0.84 & 0.84 & 1.00 & 118 & 0.37 \\
\bottomrule
\end{tabular}

\end{center}
\end{table}

\begin{table}[t!]
\caption{Llama-3-8B-Instruct (LoRA all modules): Pairwise recall of  attention heads with non-zero insight scores for SummHay Citation synthetic datasets. Recall of the real data insight heads is moderately correlated with F1 (Spearman R = 0.58).}
\label{tab:llama3_fullft_summhay_insight_recall}
\small
\begin{center}

\begin{tabular}{l||rrrrrr|rr}
\toprule
 & R,R & H,H & H,L & L,H & L,L & S,S & \# Heads & F1 \\
\midrule
R,R & 1.00 & 0.63 & 0.77 & 0.50 & 0.66 & 0.71 & 45 & 0.81 \\
H,H & 0.89 & 1.00 & 1.00 & 0.73 & 0.81 & 0.93 & 63 & 0.82 \\
H,L & 0.67 & 0.62 & 1.00 & 0.49 & 0.60 & 0.76 & 39 & 0.68 \\
L,H & 0.87 & 0.90 & 0.97 & 1.00 & 0.84 & 0.90 & 78 & 0.83 \\
L,L & 0.84 & 0.75 & 0.90 & 0.63 & 1.00 & 0.81 & 58 & 0.57 \\
S,S & 0.67 & 0.62 & 0.82 & 0.49 & 0.59 & 1.00 & 42 & 0.62 \\
\bottomrule
\end{tabular}

\end{center}
\end{table}

\subsection{Retrieval Score Cosine Similarity}

\paragraph{Across Tasks} See Table~\ref{tab:llama3_task_cosine_fullft}. Similar to fine-tuning only attention-heads, we find the highest similarity between MDQA and MuSiQue retrieval scores, and much lower similarity with SummHay Citation scores, reflecting the different nature of the task (extractive QA vs. citation).

\begin{table}[t!]
\caption{Llama-3-8B-Instruct (all LoRA modules): Cosine similarity of real dataset retrieval scores (+ SummHay insight scores) across tasks.}
\label{tab:llama3_task_cosine_fullft}
\begin{center}
\small

\begin{tabular}{l||rrrr}
\toprule
 & MDQA & MuSiQue & SummHay Retrieval & SummHay Insight \\
\midrule
MDQA & 1.00 & 0.85 & 0.35 & 0.16 \\
MuSiQue & 0.85 & 1.00 & 0.50 & 0.29 \\
SummHay Retrieval & 0.35 & 0.50 & 1.00 & 0.11 \\
SummHay Insight & 0.16 & 0.29 & 0.11 & 1.00 \\
\bottomrule
\end{tabular}

\end{center}
\end{table}

\begin{figure}[t!]
\begin{center}
\includegraphics[trim={2.5cm 0 0 0},clip,width=0.8\textwidth]{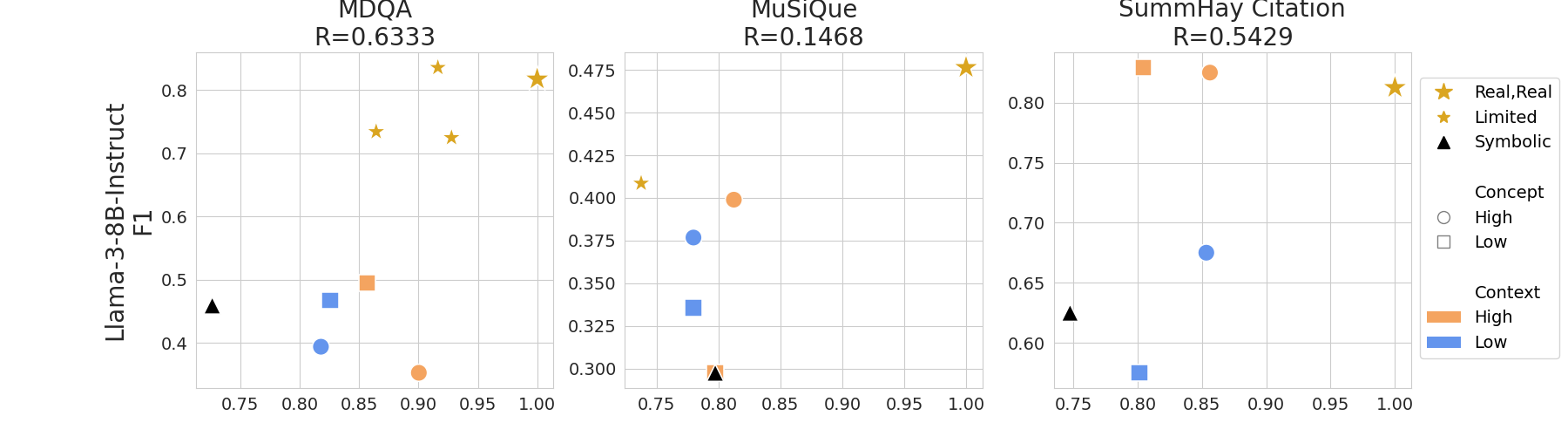}
\end{center}
\caption{Llama-3-8B-Instruct (all LoRA modules): Cosine similarity between the retrieval scores on real datasets (R, R) vs. their synthetic versions, and Spearman correlation for each setting.}
\label{fig:cosine_similarity_symbol_llama3_fullft}
\end{figure}

\paragraph{Synthetic Datasets vs. Real Task Performance} See Figure~\ref{fig:cosine_similarity_symbol_llama3_fullft}. Overall, we find that synthetic datasets with lower performance recruit fewer scoring attention heads, although the relationship is weaker than when only fine-tuning attention heads.

\subsection{Patching}
See Table~\ref{tab:llama3_patching_fullft}. Notably, we find that patching complement attention head activations is the best in more settings than patching the intersection (7 settings vs. 6 settings). This is despite the results in Table~\ref{tab:inter_compl_retrieval_scores_fullft} showing that the intersection attention heads have higher scores. 

\begin{table}[t!]
\caption{Llama-3-8B-Instruct (all LoRA modules): Results  after patching heads that comprise the complement and intersection retrieval heads between the real and synthetic data versions, compared to random retrieval heads and original performance. Best patch F1 is \textbf{bolded}, and $\Delta$ is the improvement over the original F1.}
\label{tab:llama3_patching_fullft}
\begin{center}
\small

\begin{tabular}{llllrrrrr}
\toprule
\multirow{2}{*}{Task} & \multicolumn{2}{c}{Data Variant} & \multirow{2}{*}{N}  & \multirow{2}{*}{Compl.} & \multirow{2}{*}{Inter.} & \multirow{2}{*}{Rand.} & \multirow{2}{*}{Orig.} & \multirow{2}{*}{$\Delta$} \\
 & Concept & Context &  &  &  &  &  &  \\
\midrule
\midrule
\multirow{7}{*}{MDQA} 
    & Real & Real & - & - & - & - & 0.82 & - \\
    & Real & Real (Limited) & 75 & \textbf{0.87} & 0.84 & 0.85 & 0.84 & 0.04 \\
    & Low & High & 70 & \textbf{0.66} & 0.61 & 0.56 & 0.49 & 0.17 \\
    & Low & Low & 67 & 0.61 & \textbf{0.71} & 0.44 & 0.47 & 0.24 \\
    & Symbolic & Symbolic & 72 & 0.46 & 0.33 & \textbf{0.52} & 0.46 & 0.06 \\
    & High & Low & 72 & \textbf{0.63} & 0.27 & 0.47 & 0.39 & 0.24 \\
    & High & High & 63 & 0.47 & 0.57 & \textbf{0.64} & 0.35 & 0.29 \\
\midrule
\multirow{6}{*}{MuSiQue} 
    & Real & Real & - & - & - & - & 0.48 & - \\
    & High & Low & 61 & \textbf{0.39} & 0.35 & 0.39 & 0.42 & -0.03 \\
    & Real & Real (Limited) & 59 & 0.40 & \textbf{0.42} & 0.35 & 0.41 & 0.01 \\
    & High & High & 73 & \textbf{0.41} & 0.37 & 0.31 & 0.40 & 0.01 \\
    & Low & Low & 68 & 0.39 & \textbf{0.40} & 0.35 & 0.38 & 0.02 \\
    & Symbolic & Symbolic & 51 & \textbf{0.43} & 0.10 & 0.35 & 0.37 & 0.06 \\
\midrule
\multirow{6}{*}{SummHay}
    & Real & Real & - & - & - & - & 0.81 & - \\
    & Simplified & High & 39 & 0.77 & 0.81 & \textbf{0.82} & 0.83 & -0.01 \\
    & High & High & 28 & 0.76 & 0.76 & \textbf{0.81} & 0.82 & -0.01 \\
    & High & Simplified & 24 & 0.60 & \textbf{0.72} & 0.67 & 0.68 & 0.05 \\
    & Symbolic & Symbolic & 27 & 0.64 & \textbf{0.71} & 0.66 & 0.62 & 0.08 \\
    & Simplified & Simplified & 27 & \textbf{0.64} & 0.64 & 0.61 & 0.57 & 0.07 \\
\bottomrule
\end{tabular}

\end{center}
\end{table}

\begin{table}[t!]
\caption{Llama-3-8B-Instruct (all LoRA modules): Average retrieval / insight scores for attention heads in the intersection and the complement.}
\label{tab:inter_compl_retrieval_scores_fullft}
\small
\begin{center}

\begin{tabular}{lllrr}
\toprule
\multirow{2}{*}{Task}& \multicolumn{2}{c}{Dataset Variant} & \multicolumn{2}{c}{Llama-3-8B-Instruct}\\
& Concept & Context & Inter. & Compl. \\
\midrule
\multirow{8}{*}{MDQA} 
& High & Low & 0.047 & 0.013 \\
& Real & Real (Who, When, Where) & 0.046 & 0.013 \\
& High & High & 0.049 & 0.010 \\
& Low & High & 0.045 & 0.009 \\
& Low & Low & 0.047 & 0.012 \\
& Symbolic & Symbolic & 0.049 & 0.015 \\
\midrule
\multirow{6}{*}{MuSiQue}
& Real & Real (Limited) & 0.125 & 0.049 \\
& High & High & 0.113 & 0.037  \\
& Low & High & 0.105 & 0.039  \\
& High & Low & 0.095 & 0.037 \\
& Low & Low & 0.119 & 0.045 \\
& Symbolic & Symbolic & 0.099 & 0.031 \\
\midrule
\multirow{5}{*}{SummHay}
& Simplified & Low & 0.067 & 0.010 \\
& High & Low & 0.077 & 0.020 \\
& Simplified & High & 0.065 & 0.010 \\
& High & High & 0.064 & 0.011 \\
& Symbolic & Symbolic & 0.081 & 0.013 \\
\bottomrule
\end{tabular}
\end{center}
\end{table}

\end{document}